\documentclass[lettersize,journal]{IEEEtran}
\usepackage{amsmath,amsfonts}
\usepackage{algorithmic}
\usepackage{algorithm}
\usepackage{array}
\usepackage[caption=false,font=normalsize,labelfont=sf,textfont=sf]{subfig}
\usepackage{textcomp}
\usepackage{stfloats}
\usepackage{url}
\usepackage{verbatim}
\usepackage{graphicx}
\usepackage{cite}
\usepackage{booktabs}
\usepackage{multirow}
\usepackage{amssymb}
\usepackage{textcomp}
\usepackage{color,xcolor}
\usepackage[colorlinks,linkcolor=red,citecolor=blue]{hyperref}
\begin{document}
\title{PVNet: Point-Voxel Interaction LiDAR Scene Upsampling Via Diffusion Models}

\author{Xianjing Cheng, Lintai Wu, Zuowen Wang, Junhui Hou,~\textit{Senior Member, IEEE}, Jie Wen,~\textit{Senior Member, IEEE}, Yong Xu,~\textit{Senior Member, IEEE}
\thanks{This work was supported by the Foundation of the Science and Technology project of Guangxi under Grant 
Guike AD21220114, the National Nature Science Foundation of 
China under Grants 61876051, 62466005, and  62422118,  in part by the 
Shenzhen Key Laboratory of Visual Object Detection and 
Recognition under Grant ZDSYS20190902093015527, and in part by the Hong Kong Research Grants Council under
Grants 11219324  11219422, and 11202320. 
\textit{(Corresponding author: Yong Xu, Jie Wen and Lintai Wu.)}} 
\thanks{X. Cheng is with the Telecom Guizhou Branch and also with the School of Computer Science and Technology, Harbin Institute of Technology, Shenzhen 518055, China, e-mail: chengxianjing2014@126.com.}
\thanks{L. Wu is with the College of Engineering, Huaqiao University, Quanzhou 362021, China. (email: lintaiwu2-c@my.cityu.edu.hk).}
\thanks{Z. Wang is with the School of Computer Science and Technology, Harbin Institute of Technology, 
Shenzhen 518055, China. (email: 24s151163@stu.hit.edu.cn).}
\thanks{J. Hou is with the Department of Computer Science, City
University of Hong Kong, Hong Kong SAR. (email: jh.hou@cityu.edu.hk).}
\thanks{J. Wen is with the School of Computer Science and Technology, Harbin Institute of Technology, Shenzhen 518055, China. (email: jiewen\_pr@126.com).}
\thanks{Y. Xu is with the Bio-Computing Research Center, Shenzhen Graduate
School, Harbin Institute of Technology, Shenzhen 518055, China, and also
with the Key Laboratory of Network Oriented Intelligent Computation,
Shenzhen 518055, China, email: laterfall@hit.edu.cn}  
}


\maketitle

\begin{abstract}
Accurate 3D scene understanding in outdoor environments heavily relies on high-quality point clouds. However, LiDAR-scanned data often suffer from extreme sparsity, severely hindering downstream 3D perception tasks.
Existing point cloud upsampling methods primarily focus on individual objects, thus demonstrating limited generalization capability for complex outdoor scenes. 
To address this issue, we propose PVNet, a diffusion model-based point-voxel interaction framework 
to perform LiDAR point cloud upsampling without dense supervision. Specifically, 
we adopt the classifier-free guidance-based DDPMs to guide the generation, in which we employ 
a sparse point cloud as the guiding condition and the synthesized point clouds derived from 
its nearby frames as the input. Moreover, we design a voxel completion module to 
refine and complete the coarse voxel features for enriching the feature representation. In addition, we propose a point-voxel interaction module to integrate features from both points and voxels, which efficiently improves the environmental perception capability of each upsampled point. To the best of our knowledge, our approach is the first scene-level point 
cloud upsampling method supporting arbitrary upsampling rates. Extensive experiments on various benchmarks demonstrate that our method achieves state-of-the-art performance. The source code will be available at https://github.com/chengxianjing/PVNet.
\end{abstract}


\section{Introduction}
\IEEEPARstart{A}{utonomous} driving relies on a robust perception system to accurately 
understand the surrounding environment. This system gathers external data scanned by a 
3D LiDAR sensor to perform various perception tasks \cite{wei2023agconv, wu2023leveraging,guo2024lidar, wuwssc,ma2023collaborative, c1, wu2025unsupervised, c2,  zeng2024dynamic, ren2025ddm}. 
Despite the high accuracy of LiDAR sensor positioning, the gaps between sweeps limit the 
density of point clouds. As a result, sparse point clouds hinder the perception systems from 
accurately perceiving the surrounding environment. To address this limitation, upsampling 
sparse point clouds into high-density representations has emerged as a promising solution. 
By enhancing point cloud density, the performance of various downstream tasks, such as 
navigation \cite{guo2020deep}, scene reconstruction \cite{MID, PVD, ScaleDiff}, and 
localization \cite{MSMDfusion, Point_density, Voxelnext}, can be significantly improved. \par

Current point cloud upsampling methods mainly focus on single objects and struggle to 
extend effectively to outdoor scenes \cite{PU_net, Pu-gan, Pu-gcn, Pc2, Deep-flexible, PU-flow, Pufa-gan, Dis-PU, PUGeo-Net, Deep-flexible, SpuNet, PUDM, FlattenNet,ren2023geoudf}. 
These methods consume as input sparse object-level points of a single object and 
then produce a denser point cloud. They typically rely on local geometric information 
to increase point density, using approaches like interpolation or neural network-based 
techniques. While these methods perform well on objects with simple structures, 
they often fail to densify complex and diverse outdoor environments. 
However, directly using outdoor scene data for point cloud upsampling is non-trivial. 
First, existing outdoor scene datasets lack dense point clouds for supervision. 
Second, compared to single-object point clouds, scene point clouds contain significantly 
more points, larger scales, and more complex structures.

\begin{figure*}
  \centering
  \includegraphics[width=1.0\linewidth]{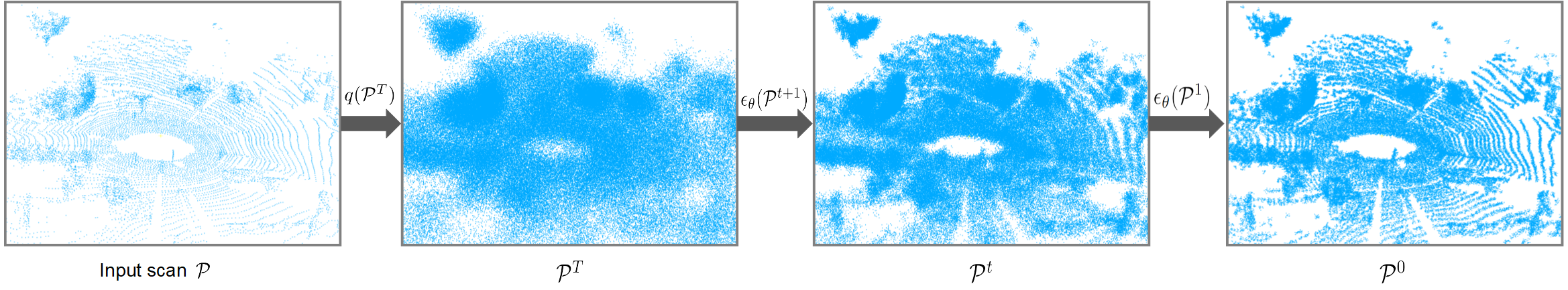}
  \caption{Visualization of our upsampling process. We start from a sparse input scan $ \mathcal{P}$ and add Gaussian noise to 
  each point denoted by $\mathcal{P}^{T}$. Next, we use our trained noise predictor $\epsilon_{\theta}$ to denoise the noisy 
  input $\mathcal{P}^{T}$ iteratively until arriving at $\mathcal{P}^{0}$, thus generating a high-quality dense LiDAR point cloud.}
  \label{fig:vis_net}
\end{figure*}

To address these issues, we propose PVNet, a \textbf{P}oint-\textbf{V}oxel Interaction LiDAR scene upsampling \textbf{Net}work based on Denoising Diffusion Probabilistic Models (DDPMs) \cite{DDPM} to generate denser scene-level point clouds. The generation process is shown in Figure \ref{fig:vis_net}. In contrast to previous 
methods that directly use high-density point clouds as ground truth, our method synthesizes 
multiple sweeps with low-density points as input and utilizes a sparse point cloud as the 
condition to guide the generation process. Specifically, 
we first add local noise to each point from the synthesized point clouds, 
and then voxelize the noisy point clouds 
to obtain the initial voxel features. 
Next, to generate more comprehensive and representative features, we design a voxel completion module to refine coarse features and recover missing voxel representations.

After capturing the distinctive voxel features, we propose a Point-Voxel Interaction Module to establish comprehensive environmental awareness for each point, enabling geometrically precise and contextually coherent upsampling in complex 3D scenes. In this module, the sparse point cloud is treated as the guiding condition to guide the generation of upsampled points. To improve the representation of each upsampled point, we interact the point with its neighboring voxels, which enhances the points' perception of their surrounding space.
Furthermore, we leverage the comprehensive features derived from guiding sparse point clouds and synthesized point clouds as weights to refine the upsampled points. 
Finally, we introduce a noise regularization term to constrain the noise distribution in the diffusion model, thereby enhancing the precision of the predictions.
We conduct extensive experiments on benchmarks to comprehensively validate the effectiveness 
of our method. 

To the best of our knowledge, our model is the first work capable of arbitrary-rate upsampling for scene-level point clouds. Besides, different from the traditional 
object-level upsampled methods that use dense ground truth to drive the learning 
process, there is no complete scene point cloud as direct supervision 
in our work. 

The key contributions of this work are as follows:
\begin{itemize}
\item We propose a novel lightweight LiDAR point cloud upsampling method based on the diffusion model. It takes the synthesized point clouds from multiple sweeps as input while the low-density point cloud is treated as the guiding condition.
\item We design a tailored voxel completion module that effectively refines coarse features and recovers missing voxel representations. This module achieves more comprehensive and representative features for complex 3D scenes.
\item We develop a point-voxel interaction module to equip each point with comprehensive environmental awareness, enabling precise point cloud upsampling.
\end{itemize}

\section{Related work}
\label{sec:releated_work}
\subsection{Point Cloud Upsampling}
Existing point cloud upsampling methods primarily focus on individual 
objects, and often fail to generalize well to complex 3D scenes. These methods typically employ   PointNet-based pipelines \cite{Pointnet} to conduct point cloud upsampling \cite{PU_net, Pu-gan, Pu-gcn, Pc2, PU-flow, Pufa-gan, Dis-PU, PUGeo-Net, Deep-flexible}.

PointNet \cite{Pointnet} leverages some shared MLps to extract individual features and then employs max pooling to generate order-invariant global features. 
PointNet++ \cite{Pointnet++} proposes a hierarchical framework to capture both local 
and global structural features, in which each level embeds a set abstraction and feature propagation units to predict upsampled points. 
PU-Net \cite{PU_net} is the first method that uses deep learning technology to perform point cloud upsampling. PUFA-GAN \cite{Pufa-gan} and PU-GAN \cite{Pu-gan} adopt the framework of the Generative Adversarial Networks (GAN) \cite{gan_goodfellow}, which expands features in the generator and utilizes a self-attention mechanism in the discriminator to enhance the dependencies between upsampled points. 
Dis-PU \cite{Dis-PU} first designs a tailored dense generator to generate upsampled points with coarse-grained details followed by a spatial refiner to minimize the 3D offsets between the predicted points and the ground truth. 
MPU \cite{Patch-PU} proposes a multi-scale architecture that progressively refines the geometric details to improve the quality of upsampled points.
PU-GCN \cite{Pu-gcn} introduces Inception DenseGCN blocks to extract distinctive geometric features followed by the NodeShuffle mechanism for feature expansion, which effectively improves the robustness of upsampled points. 
iPUNet \cite{iPUNet} employs cross fields to guide point cloud upsampling through self-supervision and uses intrinsic geometric information of the point cloud to enhance upsampling performance.
PUDM \cite{PUDM} proposes a diffusion model-based framework to achieve high-quality upsampled results at arbitrary rates. 
RepKPU \cite{rong2024repkpu} adopts a kernel-to-displacement paradigm using multi-head cross attention for point upsampling. TULIP \cite{yang2024tulip} presents a range image-based method that enhances geometric structures through a Swin-Transformer-based network. Then it converts the refined high-resolution image back to 3D points through projection. Although TULIP \cite{yang2024tulip} achieves scene-level upsampling for outdoor environments, it cannot handle arbitrary upsampling rates. To the best of our knowledge, our method is the first work capable of arbitrary-rate upsampling for scene-level point clouds.

\subsection{Denoising Diffusion Probabilistic Models}
DDPMs have recently received significant attention due to their impressive performance in the field of image generation \cite{Diffusion, DDPM, DDPM-improv, DDPM-scale, Diffusion-text, LDM, Diff-lang, Diffshift}. Furthermore, supervised diffusion training enables the guidance of generation under specified conditions and broadens them in many related generative applications \cite{ediff, classifier, adding, tang2025human}. To achieve faster generation, approximate denoising step solutions \cite{Diff-approx, Dpm-solver, lu2025dpm, DDIM} and distillation denoising model \cite{Diff-distill, Diff-distill1} are employed to address the high-time cost associated with iterative denoising steps.

Distinct from generating simple scenes, relatively few works \cite{DPM-scene, DPM-lidar, DPM-lidar1} concentrate on real-world scene generation, especially for outdoor scenes. These approaches \cite{DPM-lidar, DPM-lidar1} project 3D data onto image-based representations, enabling their generation in image domains. However, those approaches are not suitable for many real-world 3D scenes such as the projected image loses crucial depth information. Lee et al. \cite{DPM-scene} utilize a discrete diffusion model combined with a fixed voxel grid representation of the environment to achieve 3D data generation at the scene scale. 
Unlike previous work, our method fuses point-level and voxel-level features within conditional DDPMs \cite{classifier} to achieve high-quality LiDAR upsampled point clouds. 

\begin{figure*}
  \centering
  \includegraphics[width=0.9\linewidth]{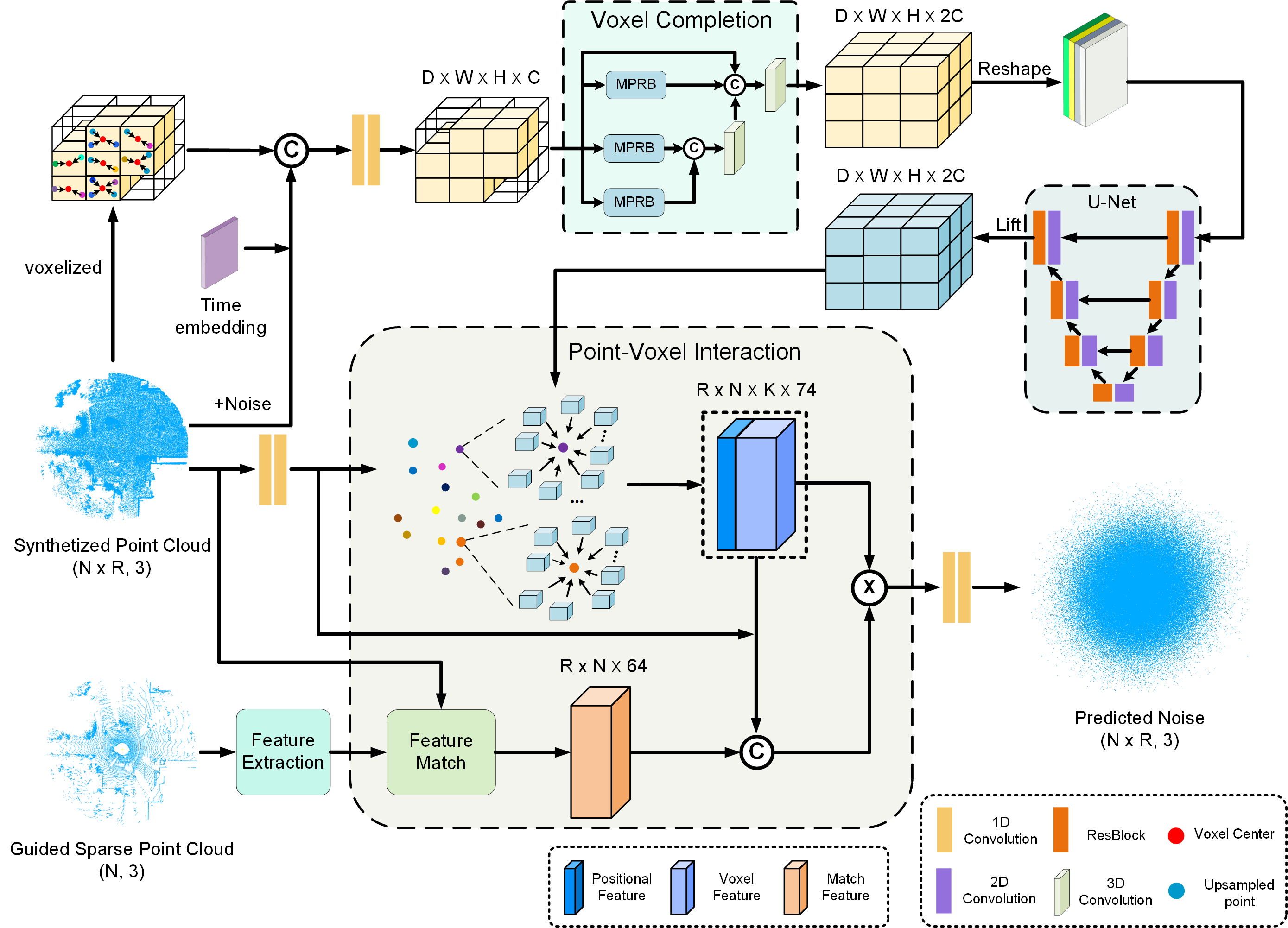}
  \caption{Flowchart of PVNet. We employ classifier-free guidance-based DDPMs to achieve the upsampling of LiDAR point clouds. In the training process, the synthesized point cloud serves as the input, while the sparse one is used as the guiding condition. We first voxelize the synthesized point cloud and propose a novel voxel completion module to enhance the voxel features. Subsequently, we adopt a U-Net module to process these features to increase their robustness. Finally, we design a tailored Point-Voxel Interaction module that efficiently integrates the features derived from the synthesized points, voxels, and sparse points to predict the noise.}
 \label{fig:framework}
\end{figure*}

\section{Proposed Method}
Given a sparse LiDAR point cloud as input, our goal is to generate a denser point cloud and achieve a more accurate representation of the surrounding scene.
Due to the absence of dense point clouds in existing LiDAR datasets, we cannot rely on ground truth to guide the learning process of the network. To address this, by using the synthesized point cloud $G$ collected from multiple consecutive sparse sweeps as input and a sparse point cloud as the condition $C$, we train a network $f$ to predict a denser upsampled point cloud $Y=f(G, C)$.

The overall architecture of PVNet is shown in Figure \ref{fig:framework}. First, we voxelize the synthesized point cloud into a voxel grid and use a two-layer MLP to produce initial voxel features embedded with diffusion time information. Then, we propose a voxel completion module comprised of three hierarchical Multi-Path ResBlocks (MPRB) to aggregate the features of occupied voxels and then diffuse them to refine and complete the features of neighboring voxels. The voxel features are further optimized through a lightweight U-Net sub-network to enhance their robustness and representation. Next, we propose a point-voxel interaction module to enhance the representation of upsampled points. This module leverages the sparse point cloud as the condition to guide the generation of dense point clouds.

\subsection{Denoising Diffusion Probabilistic Models}
\label{ddpm_theory}
DDPMs \cite{DDPM} demonstrate their effectiveness in generation tasks through an iterative denoising process, which consists of a forward diffusion process and a reverse denoising process. The diffusion process typically starts with Gaussian noise as input and iteratively adds noise into the target data over $T$ steps. The model is then trained to predict the noise added at each step $t$ and supervised by a mean squared error (MSE) loss. By comparing the predicted noise with the added noise at each step $t$, the generated data distribution gradually approximates the target distribution.

\noindent {\bf Forward Diffusion Process:}
Given a sample $x_0 \sim q(x)$ from a target data distribution, the diffusion process iteratively adds noise to $x_0$ within $T$ steps and generates a sequence $x_1$,…,$x_T$, in which $q(x_T)$ satisfies a normal distribution $N(0, I)$, where the mean is $0$ and $I$ represents a diagonal covariance matrix. 
The noise factors $\beta_1$,……,$\beta_T$ ($0 < \beta < 1$) control the intensity of noise added at each step $t$. 
At each step, noise is introduced to transform $x_{t-1}$ into $x_t$ guided by $\beta_t$. 
Typically, the noise parameter $\beta_t$ is initially defined using a linear schedule, while some studies propose alternative schedules to control the noise schedule, such as the sigmoid or cosine schedules. 
To simplify the sampling process from $x_{0}$ to $x_{t}$ given $\beta_{t}$ without explicitly computing the intermediate steps, two parameters are defined as $\alpha_t = 1 - \beta_t$ and $\Bar{\alpha}_t = \Pi_{i=1}^{t}\alpha_i$. Thus, $x_t$ is formulated as:
\begin{equation}
x_t = \sqrt{\Bar{\alpha}_t} + \sqrt{1-\Bar{\alpha}_t}\epsilon,
\label{eq:diff}
\end{equation}
where $\epsilon \sim N(0, I)$.

\begin{figure}[t]
  \centering
   \includegraphics[width=1.0\linewidth]{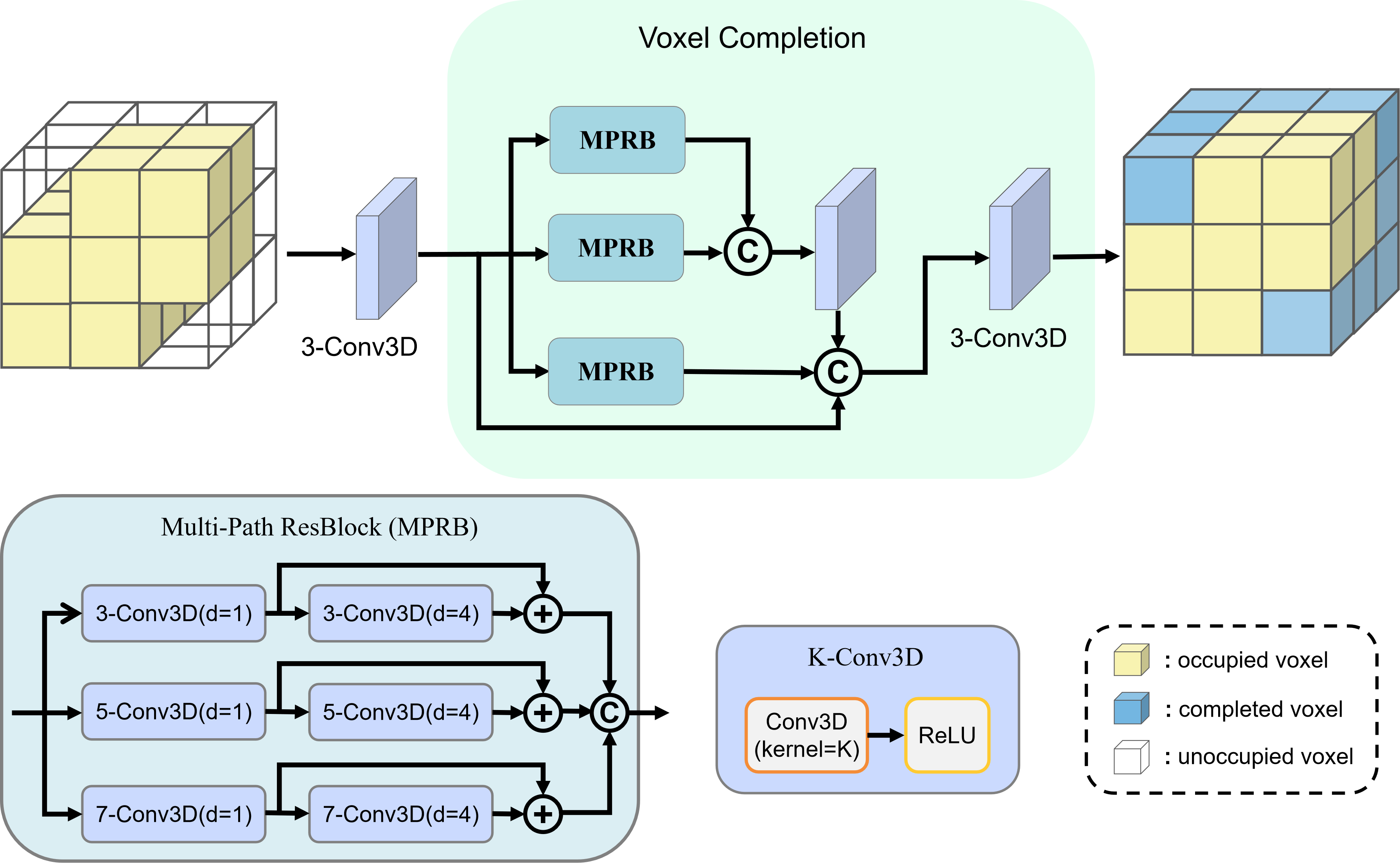}
   \caption{Overview of the voxel completion module. This module employs three hierarchical Multi-Path ResBlocks that progressively aggregate the features of occupied voxels and diffuse them to neighboring voxels to refine and complete the features. Conv3D: 3D convolution, d: dilation rate.}
   \label{fig:completion}
\end{figure}

\noindent {\bf Reverse Denoising Process:}
The reverse denoising process predicts the added noise $\epsilon$ at each step $t$ and uses it to iteratively reconstruct $x_0$ from $x_T$. Starting from an initial noise sample $x_{T}$, the reverse process gradually denoises the sample until reaching $x_{0}$, which is the generated sample. 
The denoising process is defined as follows:
\begin{equation}
\label{eq:reverse}
x_{t-1} = x_t - \frac{1-\alpha_t}{\sqrt{1-\Bar{\alpha}_t}}\epsilon_\theta(x_t,t) + \frac{1-\Bar{\alpha}_{t-1}}{1-\alpha_t}\beta_t \mathcal{N}(0,\mathit{I}),
\end{equation}
where $\epsilon_\theta(x_t, t)$ is the predicted noise. The process of recovering the original data distribution from the noisy distribution can be formulated as a Markov process. Moreover, the denoising process can be further guided by a specific condition $\mathit{C}$, which is called classifier-free guidance \cite{classifier}. In this work, we use classifier-free guidance-based DDPMs to achieve high-quality point cloud upsampling. 

\noindent {\bf Classifier-free Guidance:}
Given an initial sample $x_0$, a condition $c$ and a random step $t \in [0, T]$, the model utilizes an $\mathcal{L}_2$ loss to drive the training process, which is defined as follows:
\begin{equation}
\mathcal{L}(x_t, c, t) = ||\epsilon - \epsilon_{\theta}(x_t, c, t)||^{2}.
\label{eq:L2}
\end{equation}

During inference, classifier-free guidance \cite{classifier} starts from $x_{T} \in N(0, I)$ and iteratively denoises it until reaching $x_0$. The predicted noise is a combination of the conditional and unconditional noise distributions, which is computed as follows:
\begin{equation}
\epsilon'_{\theta}(x_t, c, t) = \epsilon_{\theta}(x_t, \emptyset, t) +s[\epsilon_{\theta}(x_t, c, t) - \epsilon_{\theta}(x_t, \emptyset, t)],
\label{eq:result}
\end{equation}
where $s$ is a weighting factor, $\epsilon_{\theta}(x_t, \emptyset, t)$ represents the unconditional noise prediction. 
We use Eq. (\ref{eq:result}) to compute the noise at each step $t$ and incorporate Eq. (\ref{eq:reverse}) to compute $x_{T-1}$ until generating a new sample $x_0$. The final output $x_0$ is a generated sample conditioned on $c$.

\subsection{Voxel Completion Module}
In the forward diffusion process, we take synthesized point clouds as input. First, we voxelize the points into a grid of size $128 \times 128 \times 16$. Subsequently, we compute the offset between each point and its corresponding voxel center, which serves as the feature representation of the voxelized points.

Meanwhile, we add noise to each point, which is defined as: 
\begin{equation}
p_t = p + \sqrt{1-\Bar{\alpha}_t}\epsilon_{t},
\label{eq:define_pt}
\end{equation}
where $p$ represents the 3D coordinate of the point, $\epsilon_{t}$ is a random noise following a normal distribution at step $t$, and the derivation of $\Bar{\alpha}_t$ is provided in Section \ref{ddpm_theory}. Note that we add noise to the $x$, $y$, and $z$ coordinates of each point, which means we change the positions of existing points rather than adding more points.

After generating the local noisy points, we integrate the temporal information with the features derived from both the voxelized points and noisy points. These concatenated features are then fed into a two-layer MLP to produce the initial voxel-based features. 
We observe that voxel features processed solely through MLPs tend to be overly coarse, with many voxels exhibiting inadequate feature representation or even missing features in occupied regions. To address this limitation, we propose a voxel completion module that not only refines the voxel features but also propagates features from occupied voxels to their neighboring voxels with missing features, thereby enhancing the overall feature representation capability.

The pipeline of the voxel completion module is detailed in Figure \ref{fig:completion}. We first use a small-stride 3D convolution block to extract fine-grained local features. Following this, we design three hierarchical Multi-Path ResBlock (MPRB) units to facilitate progressive refinement of the features. In each MPRB, we propose three parallel 3D convolutions with different kernel sizes and dilation rates, and we incorporate a residual connection at each level to enhance the representation of voxel features. 
Three groups of features with different receptive fields are then concatenated to increase their diversity and robustness. 
Finally, we integrate the features from three MPRB units with the initial features to refine and complete the features of neighboring voxels. This process is also particularly beneficial in improving the representation of occluded regions. Then, the voxel features are reshaped and refined by a 2D U-Net sub-network from LMSCNet \cite{LMSCNet} to further enhance their robustness.

Note that we remove the widely used batch normalization layer and disable the bias in the 3D convolutions of our voxel completion module to preserve the spatial relations between voxels and maintain the structural integrity within voxels. Batch normalization and bias have been widely used in previous 3D scene reconstruction methods, and there has been no research investigating their negative effects. 
In fact, since the range of scene point clouds varies significantly across different voxels, normalization layers such as Batch Normalization would rescale the distribution of voxel features, which potentially disrupts the spatial relations between voxels. 
Furthermore, the bias term in 3D convolution can alter the original structures of voxels, which potentially distorts the structural integrity within voxels. 
Thus, we remove all the normalization layers and disable the bias in 3D convolution to preserve the two spatial properties of voxels during optimization. Detailed analysis of these theories is demonstrated in the ablation studies. 

\begin{figure*}[t]
  \centering
   \includegraphics[width=1.0\linewidth]{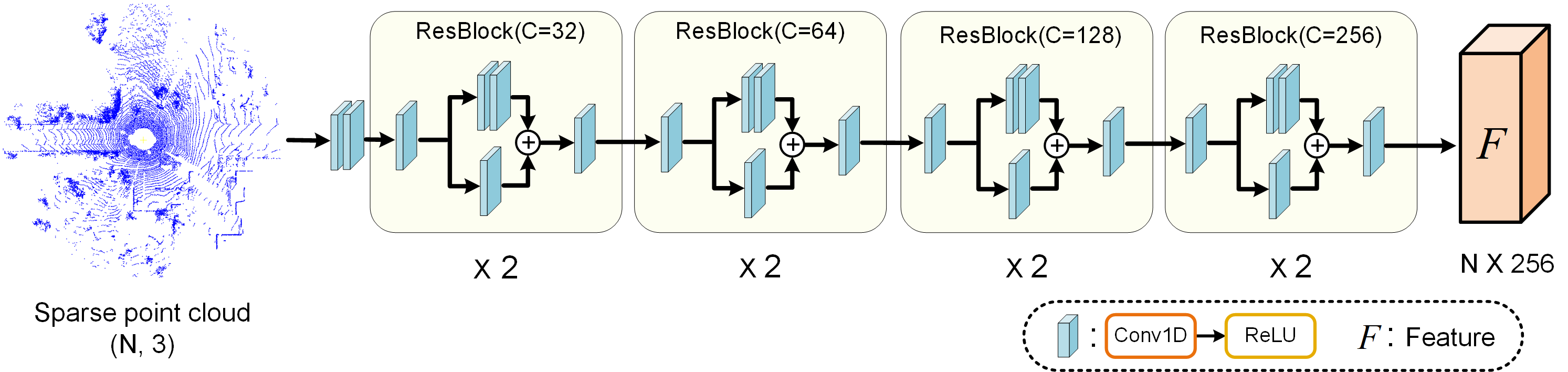}
   \caption{Pipeline of the feature extraction module. This module takes a sparse point cloud as input and employs 4 consecutive ResBlocks with increasing channels to extract the high-dimension features of the sparse point cloud. $F$ represents the feature of the sparse point cloud.}
   \label{fig:feat_part}
\end{figure*}

\subsection{Point-Voxel Interaction Module}
\label{sec:PVI_module}
Unlike single-object point clouds, real-world scenes involve multiple objects with complex spatial relationships and diverse backgrounds. To achieve high-quality upsampling, each point is required to accurately perceive and integrate information from its surroundings to capture both local geometric structures and global scene semantics. However, point-based methods struggle to model long-range dependencies efficiently due to high computational costs, while purely voxel-based representations lose fine-grained geometric details.  To address these challenges, we propose a Point-Voxel Interaction Module that enables each point to efficiently perceive its surrounding contextual information, thereby achieving more accurate upsampling.

Based on the indices of non-empty voxels, we compute the world coordinate of these voxels using their actual sizes. Then, we perform the interaction between the upsampled points and the voxels to enhance the perception of the upsampled points with respect to their surroundings. Specifically, each upsampled point searches 16 neighboring voxels as a group using the KNN algorithm \cite{KNN}. We term the features of the neighboring voxels in all groups as $F_{group}$. Note that the number of voxels (128 $\times$ 128 $\times$ 16) should be significantly greater than the number of input points to ensure that each point has a rich representation. Next, within each group, the upsampled point interacts with its 16 neighboring voxels based on their positional relation. This interaction encompasses the coordinates from the point and its neighboring voxels, the offsets between the point to each voxel, and the total distances between the point and the voxels. 
The definition of these positional features $F_{pos} \in \mathbb{R}^{N \times 16 \times 10}$ in a group is as follows:
\begin{equation}
F_{pos} = \texttt{Cat}(P_{v}, P_{p}, \Delta_{v2p}, ||\Delta_{v2p}||_2^2),
\label{eq:position_feat}
\end{equation}
where $P_{v}, P_{p}$ are the world coordinates of the center point and the neighboring voxels, respectively. $\Delta_{v2p} \in \mathbb{R}^{N \times 16 \times 3}$ is the offsets between the upsampled point and each voxel within a group, $||\Delta||_{L2} \in \mathbb{R}^{N \times 16 \times 1}$ is the L2-norm distance of $\Delta$. 
Besides, in this module, we use a sparse point cloud as the guiding condition to generate the corresponding upsampled point cloud. To enable interaction with the guiding point cloud, we propose a feature extraction module to extract the features of the sparse point cloud. 

\noindent {\bf Interaction with the guiding sparse point cloud:} In this work, we use the classifier-free guidance-based DDPMs to achieve scene-level point cloud upsampling. Thus, we use a sparse point cloud as the guiding condition and design a feature extraction module to extract distinctive features from the guiding point cloud. The details of this module are shown in Figure \ref{fig:feat_part}. We adopt 4 consecutive ResBlocks with varying channel sizes to generate robust and distinctive guiding features. 

In the classifier-free guidance-based DDPMs, it is crucial to establish a strong connection between the features from the guiding sparse points and the input synthesized points. To achieve this, we find the nearest sparse points for each input point by computing the minimum Euclidean distance. We then use the indices of the nearest sparse points to retrieve the most relevant features from the synthesized points, which are termed as the match features between the input points and the sparse points. The match features are effectively utilized to enhance the relation between the input points and sparse points. Finally, we employ the shallow MPLs to refine the match features, and the result is denoted by $F_{match} \in \mathbb{R}^{N \times 64}$.

In the next step, we concatenate the related features, producing the hybrid feature $F_{hybr}$:
\begin{equation}
F_{hybr} = \texttt{Cat}(F_{pos}, F_{group}, F_{points}, F_{match}),
\label{eq:related_feat}
\end{equation}
where  $F_{points}$ is the features of upsampled points optimized by two-layer 1D convolutions. Then, we use a two-layer 1D convolution block to optimize $F_{hybr}$ as the refined weight $W_{F_{hybr}}$. The enhanced features within groups are averaged and passed through two 1D convolutions to generate the final upsampled point cloud $P_{up}$. This procedure is detailed as follows: 
\begin{equation}
P_{up} = \texttt{Average}(W_{F_{hybr}} \times F_{group}).
\label{eq:upsample_pts}
\end{equation}

\begin{figure}[t]
  \centering
   \includegraphics[width=1.0\linewidth]{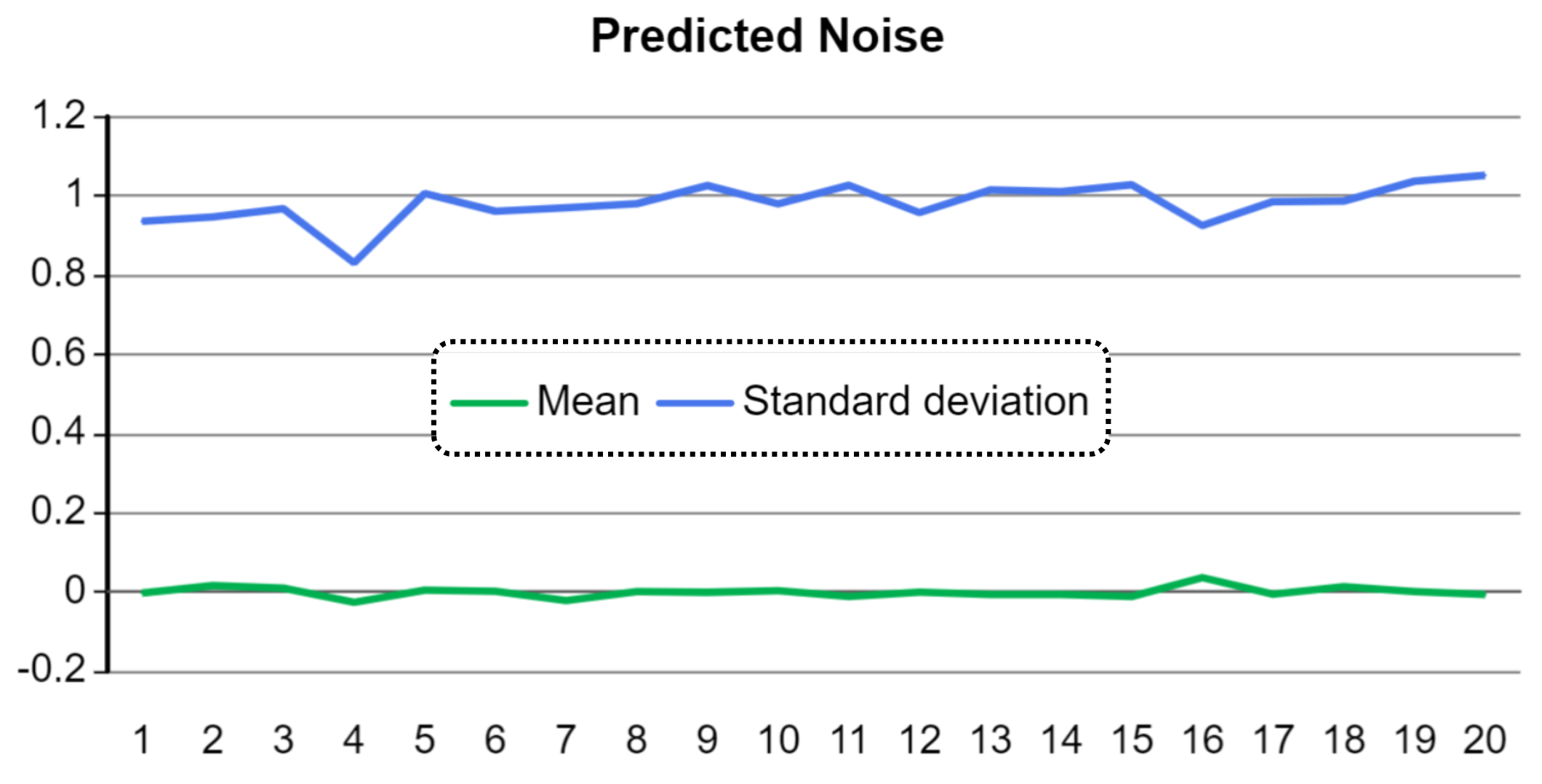}
   \caption{Mean and standard deviation of the predicted noise without the noise regularization.}
   \label{fig:distri_noise}
\end{figure}

\subsection{Noise Regularization}
\label{sec:noise_regu}
The traditional DDPMs use a leveraged formulation\cite{ScaleDiff} as Eq. \eqref{eq:L2} to predict the noise added to the original data. This classified formulation is optimized using an $\mathcal{L}_2$ loss between the prediction and the added noise, in which the predicted noise follows a normal distribution, i.e., $\epsilon_{\theta} \sim \mathcal{N}(0, I)$. However, since we add the noise to each point in the world coordinate system formulated by Eq. \eqref{eq:define_pt}, the prediction of our model during training does not follow a standard normal distribution, as shown in Figure \ref{fig:distri_noise}. As the denoising process progresses, the mean of the predicted noise approaches zero, but its standard deviation fluctuates significantly and deviates from 1. Thus, to enable the predicted noise $\epsilon_{\theta} \approx \mathcal{N}(0, I)$, we propose a noise regularization term based on the $\mathcal{L}_2$ loss to further improve the quality of the upsampled point cloud. Due to the superior performance of the smooth $L_1$ loss function in the deep learning field, we choose it as our regularization function, which is formulated as follows:
\begin{equation}
    \mathcal{L}_{std} = smooth_{L1} (\epsilon_{\theta} - 1),
\label{eq:regularization}
\end{equation}
in which
\begin{equation}
    smooth_{L1} = \left\{ \begin{array}{rcl}
         & 0.5x^2, & \mbox{if} |x|<1  \\
         & |x|-0.5, & \mbox{otherwise}
    \end{array}\right.
\end{equation}

In total, our final training loss is defined as follows:
\begin{equation}
    \mathcal{L} = ||\epsilon_{\theta} - \epsilon||_{2}^2 + \lambda\mathcal{L}_{std},
\label{eq:loss_total}
\end{equation}
where $\lambda$ is a weighting coefficient used to balance the effect of the regularization weight.

\section{Experiments}
\label{sec:experiment}
\subsection{Experimental Setting}
\label{sec:experiment_set}
\textbf{Datasets.} In this work, we use two outdoor datasets, i.e., SemanticKITTI \cite{SemanticKITTI1, KITTI} and KITTI-360 \cite{liao2022kitti}, to comprehensively assess the performance of our method. SemanticKITTI provides a large number of outdoor LiDAR scans with point-wise annotations from KITTI Odometry Benchmark \cite{KITTI, KITTI2015}. The dataset is divided into training/validation/test sets, which comprise 10/1/10 sequences, respectively. We sample 3834 scans from the training sequences of SemanticKITTI as training data and 40/200 scans from sequence 08 as validation/test data. KITTI-360 \cite{liao2022kitti} is a large-scale outdoor dataset that contains 320k images and 100k laser scans in a driving distance of 73.7 km, and provides a full 360\textdegree field of view equipped with the additional cameras and the pushbroom laser scanner. Following the same evaluation protocol as SemanticKITTI, we also use 200 scans in KITTI360 as test data.

\textbf{Implementation details.} We train PVNet for 20 epochs with a batch size of 2 on two NVIDIA Tesla A40 GPUs. We use the Adam optimizer \cite{Adam} with an initial learning rate of $10^{-4}$, which is decreased by half every 5 epochs. The weight decay is set to $10^{-4}$. The upsampling rate $R=10$ during the training stage, while the upsampling rate can be arbitrary during the inference stage. In the forward diffusion process, we adopt a linear noise schedule where the noise levels increase from $\beta_{0}=3.5\times10^{-5}$ to $\beta_T=7\times10^{-3}$ over $T=1000$ steps.
We adopt the strategy of ScalingDiff \cite{ScaleDiff} to generate the synthesized point clouds. Specifically, for each sequence, we first align the multi-frame LiDAR scans using their pose parameters. Then, we merge the aligned scans and select 8 complete and dense point cloud maps, while removing moving objects and artifact points. We extract the corresponding synthetic point cloud from the complete map based on the intrinsic parameters of the LiDAR scan as our input.
The point number of the guiding sparse point cloud is set to 18000 and the synthesized point cloud contains 180000 points. The size of voxel grids is $128 \times 128 \times 16$ and the parameter $\lambda$ is 1.0 in all the experiments. During inference, we employ DPMSolver \cite{Dpm-solver} to predict the dense upsampled point clouds with the number of denoising steps reduced to 50.

\textbf{Evaluation Metrics.} SemanticKITTI \cite{SemanticKITTI1} does not provide the dense 
ground truth to evaluate the performance in the LiDAR point cloud upsampling task. 
In this case, we leverage two evaluation metrics that do not require ground truth to 
evaluate our method as comprehensively as possible. First, since the shape of the 
upsampled result needs to match the shape of the input sparse point cloud, we use 
Chamfer Distance (CD) to compute the distance between the upsampled point cloud and the 
sparse input. 
A large CD value indicates that the upsampled result deviates from the input shape while a small CD value suggests that the upsampled shape is complete and accurate. Note that when the model's output is identical to the original input, the CD value is 0. In this case, the model does not possess upsampling capability, but this situation can be identified through visual results.
The formulation of CD is defined as follows:
\begin{equation} 
\begin{split}
  CD(P,Q)=\frac{1}{|P|}\sum\limits_{p \in P} \underset{q \in Q}{min}|| p-q||^{2}_{2} \\ + \frac{1}{|Q|}\sum\limits_{q \in Q} \underset{p\in P}{min}||p-q||^{2}_{2},
\label{eq:offset_CD}
\end{split}
\end{equation}
where $P$ is the upsampled dense points, $Q$ is the input sparse points.

\begin{figure}
  \centering
   \includegraphics[width=1.0\linewidth]{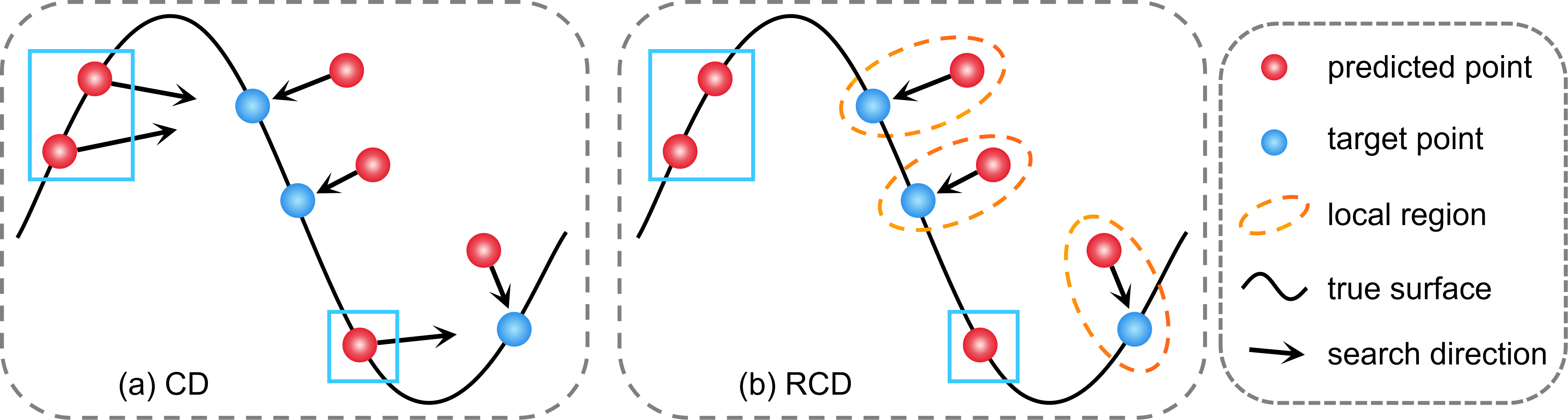}
   \caption{Comparison between CD and RCD metrics. (a) CD measures the nearest-neighbor distance for every point between the input and predicted points, (b) RCD evaluates point distances in local regions.}
\label{fig:metric}
\end{figure}

Although the CD metric can assess the completeness and precision of the overall shape of the upsampled point cloud, it is less effective in assessing fine-grained local structures, which are also critical for scene point cloud upsampling. To better evaluate the performance, we introduce the Region-aware Chamfer Distance (RCD) metric proposed by P2C \cite{P2C} to measure the similarity between the local regions of the input and the predicted point. As shown in Figure \ref{fig:metric}, when computing CD, the predicted point searches for the nearest neighbors in the whole point cloud. On the contrary, RCD evaluates the point distances in local regions. Specifically, we use FPS \cite{FPS} to divide the target points into several groups, and within each group (i.e., local region), each predicted point searches for the nearest target point to compute its distances as described in Figure {\ref{fig:metric}} (b).
The formulation is defined as follows:
\begin{equation} 
\begin{split}
 RCD(P,Q)=\frac{1}{|R_P|}\sum\limits_{p \in R_P} \underset{q \in R_Q}{min}|| p-q||_{2} \\ + \frac{1}{|R_Q|}\sum\limits_{q \in R_Q} \underset{p\in R_P}{min}||p-q||_{2},
\label{eq:RCD}
\end{split}
\end{equation}
where 
\begin{equation} \nonumber
\begin{aligned}
  & R_P = \bigcup_{i=1}^{n}\{N_{k}^{P}(p_i)|p_i \in P\} \\
  & R_Q = \bigcup_{j=1}^{m}\{N_{k}^{Q}(q_j)|q_j \in Q\},
\label{eq:region}
\end{aligned}
\end{equation}
where $n,m$ are the number of the upsampled points $P$ and input points $Q$ in local regions, respectively. $N_k$ is the $k$-nearest neighbors. We set the number of groups to 64 and each group contains 32 target points. 
To further evaluate the robustness of the network, we randomly split the total groups of RCD into two subsets: Recon-RCD (20 groups) and Match-RCD (44 groups). This aims to assess the shape similarities between predicted points and target points across different scales of local regions.

\renewcommand{\arraystretch}{1.3}
\begin{table*}
  \centering
  \fontsize{9}{9}\selectfont
  \caption{Quantitative comparison between PVNet and state-of-the-art approaches on SemanticKITTI dataset. $R$ represents the upsampling rate.}
  \label{tab:comparison_benchmarks}
  \setlength{\tabcolsep}{1.2mm}{
      \begin{tabular}{cccccc@{\hspace{10mm}}cccccccc}
        \toprule
        \multirow{2}{*}{Methods} &\multicolumn{5}{c}{$\times$ 4 (R=4)} &\multicolumn{5}{c}{$\times$ 16 (R=16)} \\
        \cmidrule(r{10mm}){2-6}  \cmidrule(r){7-11}  & CD $\downarrow$ & RCD $\downarrow$ & Recon-RCD $\downarrow$ & Match-RCD $\downarrow$ & Time(s) $\downarrow$ & CD $\downarrow$   &RCD $\downarrow$  &Recon-RCD $\downarrow$  & Match-RCD $\downarrow$ & Time(s) $\downarrow$ \\
        \midrule
        PU-Net \cite{PU_net} & 0.769 & 0.696 & 0.707 & 0.703 & 1.456 & 1.836 & 1.022 & 1.065 & 1.038  & \textbf{18.515} \\
        PU-GAN \cite{Pu-gan} & 0.353 & 0.533 & 0.565 & 0.547 & 10.898 & 0.645 & 0.708 & 0.767 & 0.724  & 339.549 \\
        Dis-PU \cite{Dis-PU} & 0.125 & 0.432 & 0.467 & 0.446 & 14.192 & 0.164 & 0.706 & 0.760 & 0.729  & 230.782\\
        PU-GCN \cite{Pu-gcn} & 0.112 & 0.431 & 0.466 & 0.445 & 11.302 & 0.150 & 0.687 & 0.770 & 0.714  & 217.434\\
        PUDM \cite{PUDM}     & 614.4 & 16.31 & 17.42 & 16.67 & 11.564 & 614.5 & 16.31 & 17.59 & 16.59  & 92.928 \\
        RepKPU \cite{rong2024repkpu} & 0.118 & 0.418 & 0.458 & 0.426 & 10.453 & 0.153 & 0.698 & 0.762 & 0.716 & 215.969 \\
        TULIP \cite{yang2024tulip} & 12.949 & 0.959 & 1.154 & 1.033 & \textbf{0.389} & - &- & - &- &-\\
        \textbf{PVNet(Ours)} & \textbf{0.097} & \textbf{0.342}  & \textbf{0.365} & \textbf{0.348}  & 43.13
        & \textbf{0.082} & \textbf{0.655} &\textbf{0.706} & \textbf{0.675}  & 84.22\\
        \bottomrule
      \end{tabular}
      }
\end{table*}

\renewcommand{\arraystretch}{1.3}
\begin{table*}
  \centering
  \fontsize{9}{9}\selectfont
  \caption{Quantitative comparison between PVNet and state-of-the-art approaches on KITTI-360 dataset. $R$ represents the upsampling rate.}
  \label{tab:comparison_KITTI-360}
  \setlength{\tabcolsep}{1.2mm}{
      \begin{tabular}{cccccc@{\hspace{10mm}}cccccccc}
        \toprule
        \multirow{2}{*}{Methods} &\multicolumn{5}{c}{$\times$ 4 (R=4)} &\multicolumn{5}{c}{$\times$ 16 (R=16)} \\
        \cmidrule(r{10mm}){2-6}  \cmidrule(r){7-11}  & CD $\downarrow$ & RCD $\downarrow$ & Recon-RCD $\downarrow$ & Match-RCD $\downarrow$ & Time(s) $\downarrow$ & CD $\downarrow$   &RCD $\downarrow$  &Recon-RCD $\downarrow$  & Match-RCD $\downarrow$ & Time(s) $\downarrow$ \\
        \midrule
        PU-Net \cite{PU_net} & 1.727 & 1.343 & 1.401 & 1.359 & 2.695 & 3.290 & 2.037 & 2.216 & 2.073  & \textbf{41.257} \\
        PU-GAN \cite{Pu-gan} & 1.778 & 1.079 & 1.218 & 1.128 & 16.840 & 3.989 & 1.389 & 1.607 & 1.475  & 303.117 \\
        Dis-PU \cite{Dis-PU} & 0.355 & 0.882 & 1.009 & 0.939 & 14.103 & 0.507 & 1.340 & 1.580 & 1.426  & 231.293\\
        PU-GCN \cite{Pu-gcn} & 0.321  &  0.862 & 0.991  &  0.928 & 11.288  & 0.443  &  1.318 & 1.569  &  1.410  & 217.089 \\
        PUDM \cite{PUDM}     & 1095.1  & 26.47  & 29.70  & 27.34  & 44.593  & 1095.2  & 26.47  & 29.59  &  27.51  & 196.824 \\
        RepKPU \cite{rong2024repkpu} & 0.353  & 0.847  & 0.966  & 0.903  & 18.158  & 0.493  &  1.319 & 1.530  &  1.419  & 219.577 \\
        TULIP \cite{yang2024tulip} & 9.330 & 2.010  & 2.458  &  2.198 & \textbf{0.374}  & - &- & - &- &-\\
        \textbf{PVNet(Ours)} & \textbf{0.173} & \textbf{0.578}  & \textbf{0.638} & \textbf{0.602}  & 42.98
        & \textbf{0.148} & \textbf{1.184} &\textbf{1.389} & \textbf{1.261}  & 78.06\\
        \bottomrule
      \end{tabular}
      }
\end{table*}

\begin{figure*}
  \centering
    \begin{tabular}{@{}c@{}c@{}c@{}c}
      \begin{tabular}{@{}c}
        \rotatebox{90}{(a) Input}
      \end{tabular}
      &
      \begin{tabular}{@{}c}
        \includegraphics[width=0.35\textwidth]{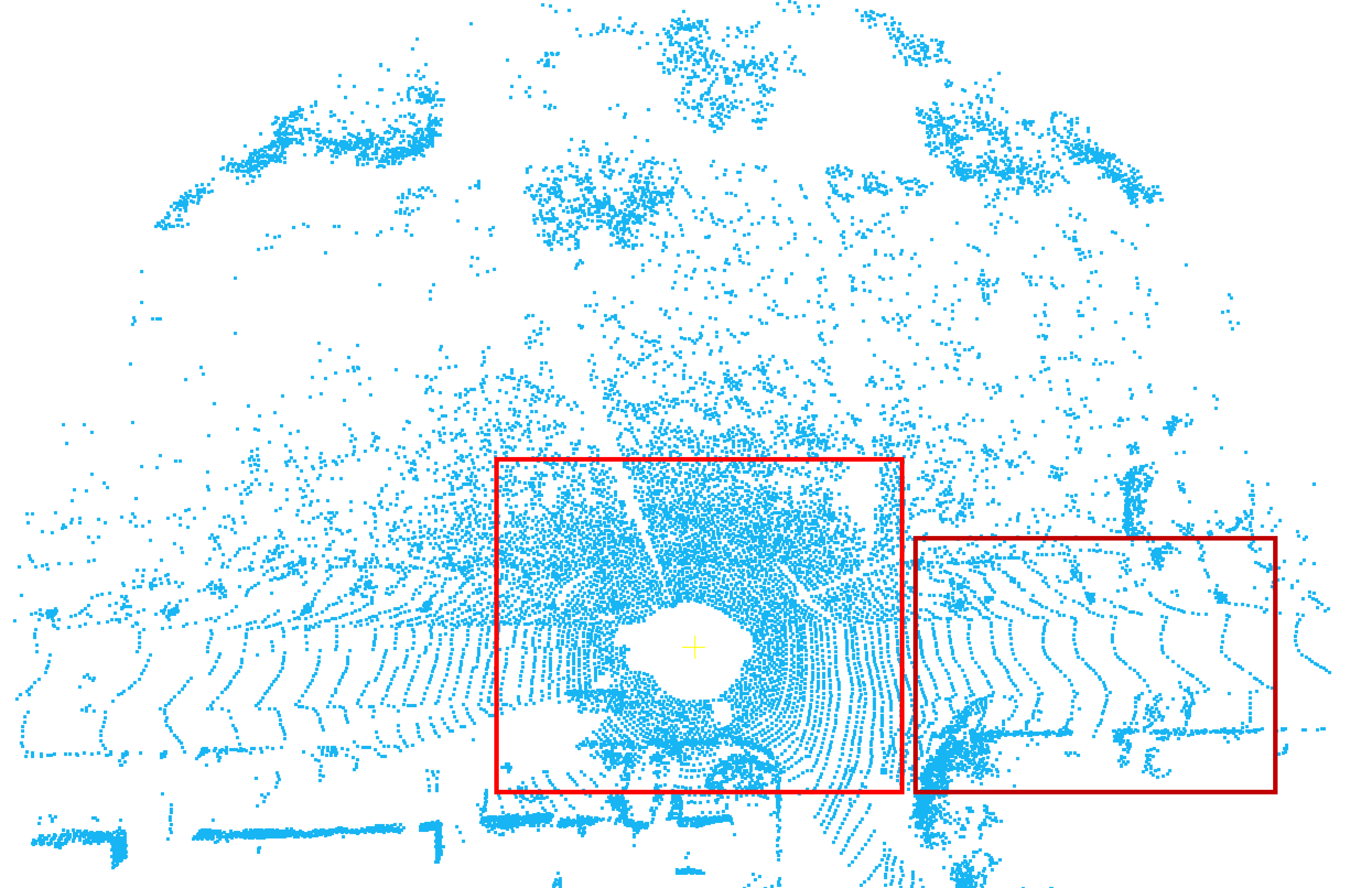}
      \end{tabular}
      &
      \begin{tabular}{@{}c}
        \includegraphics[width=0.25\textwidth]{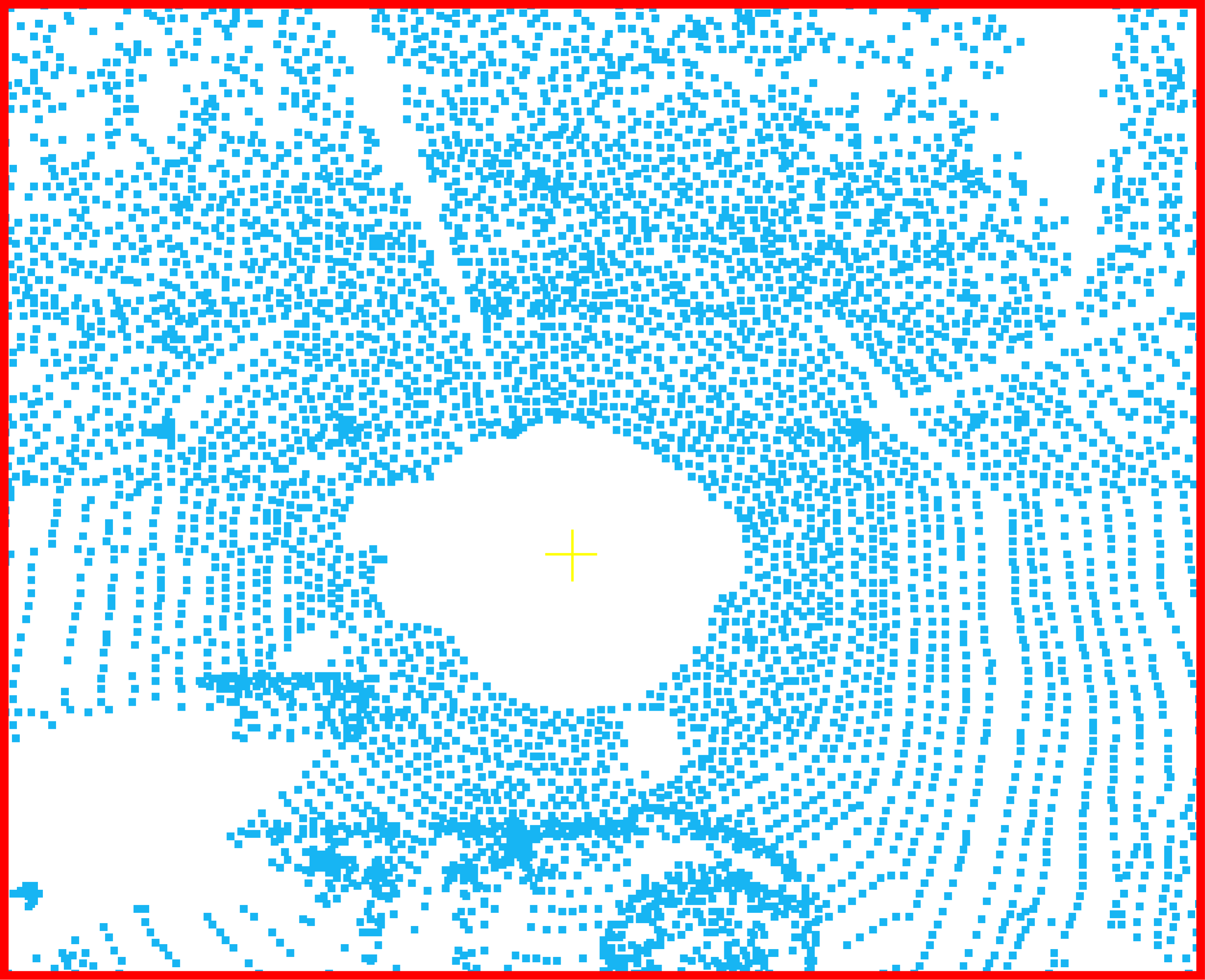}
      \end{tabular}
      &
      \begin{tabular}{@{}c}
        \includegraphics[width=0.3\textwidth]{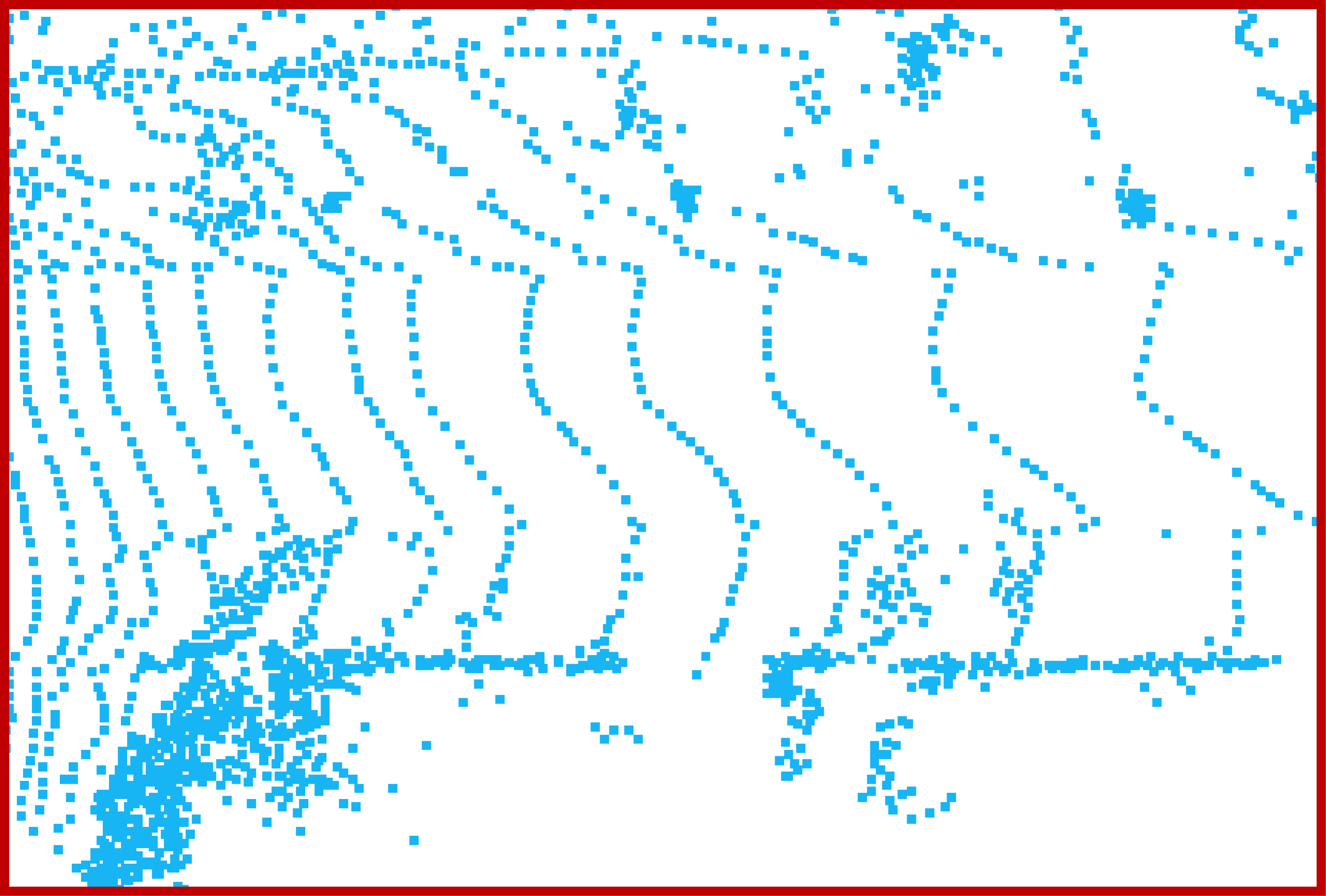}
      \end{tabular}
      
      \\
      \begin{tabular}{@{}c}
        \rotatebox{90}{(c) PU-GCN \cite{Pu-gcn}}
      \end{tabular}
      &
      \begin{tabular}{@{}c}
        \includegraphics[width=0.35\textwidth]{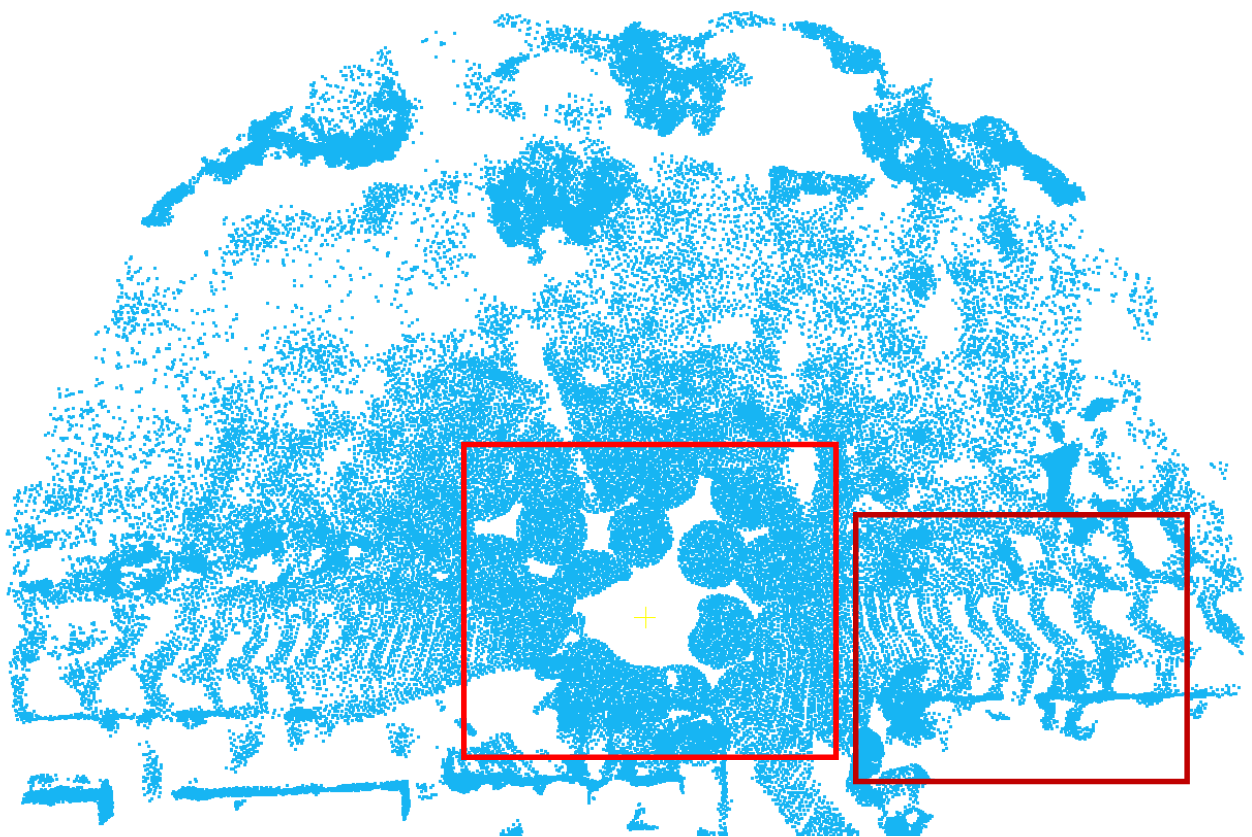}
      \end{tabular}
      &
      \begin{tabular}{@{}c}
        \includegraphics[width=0.25\textwidth]{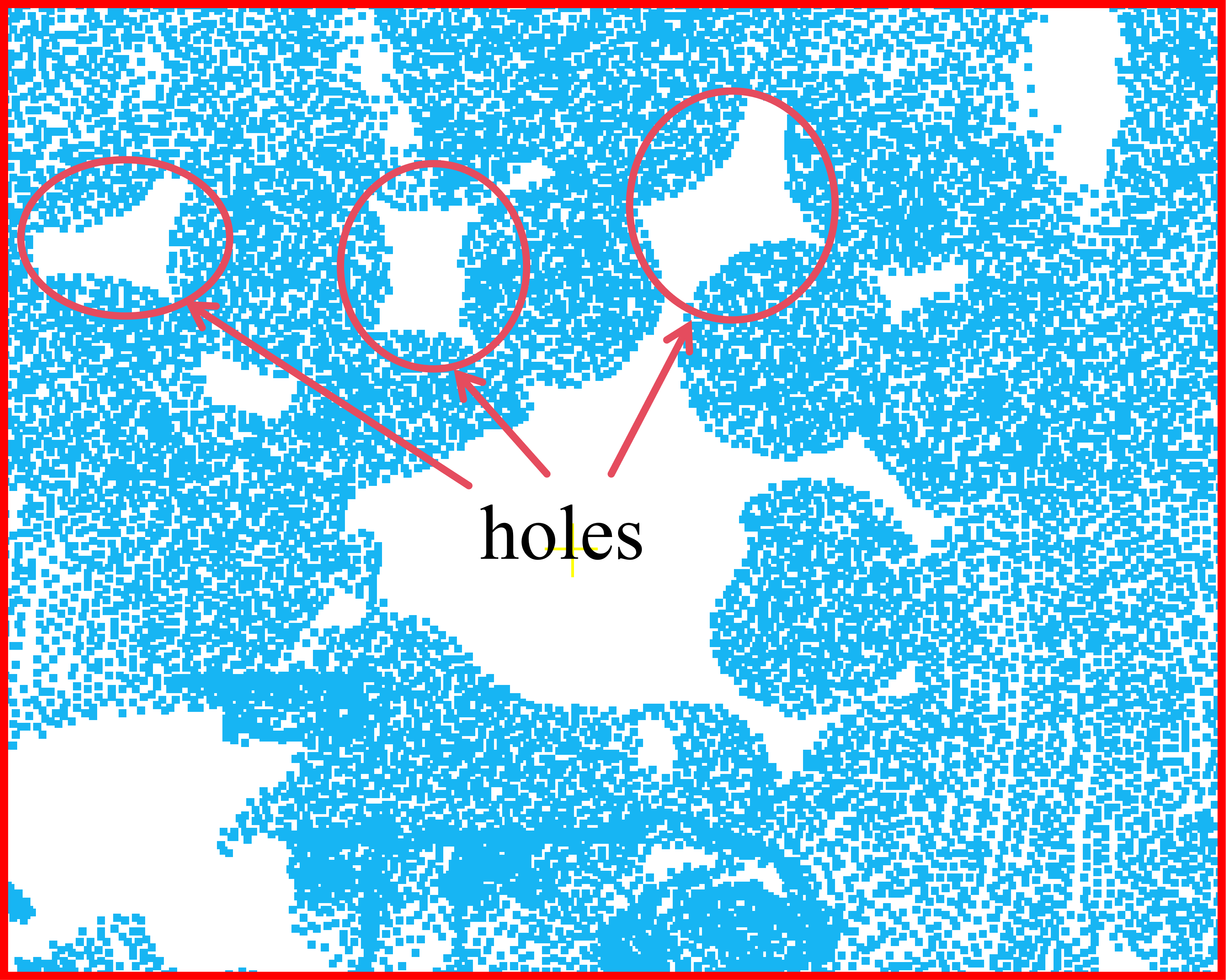}
      \end{tabular}
      &
      \begin{tabular}{@{}c}
        \includegraphics[width=0.3\textwidth]{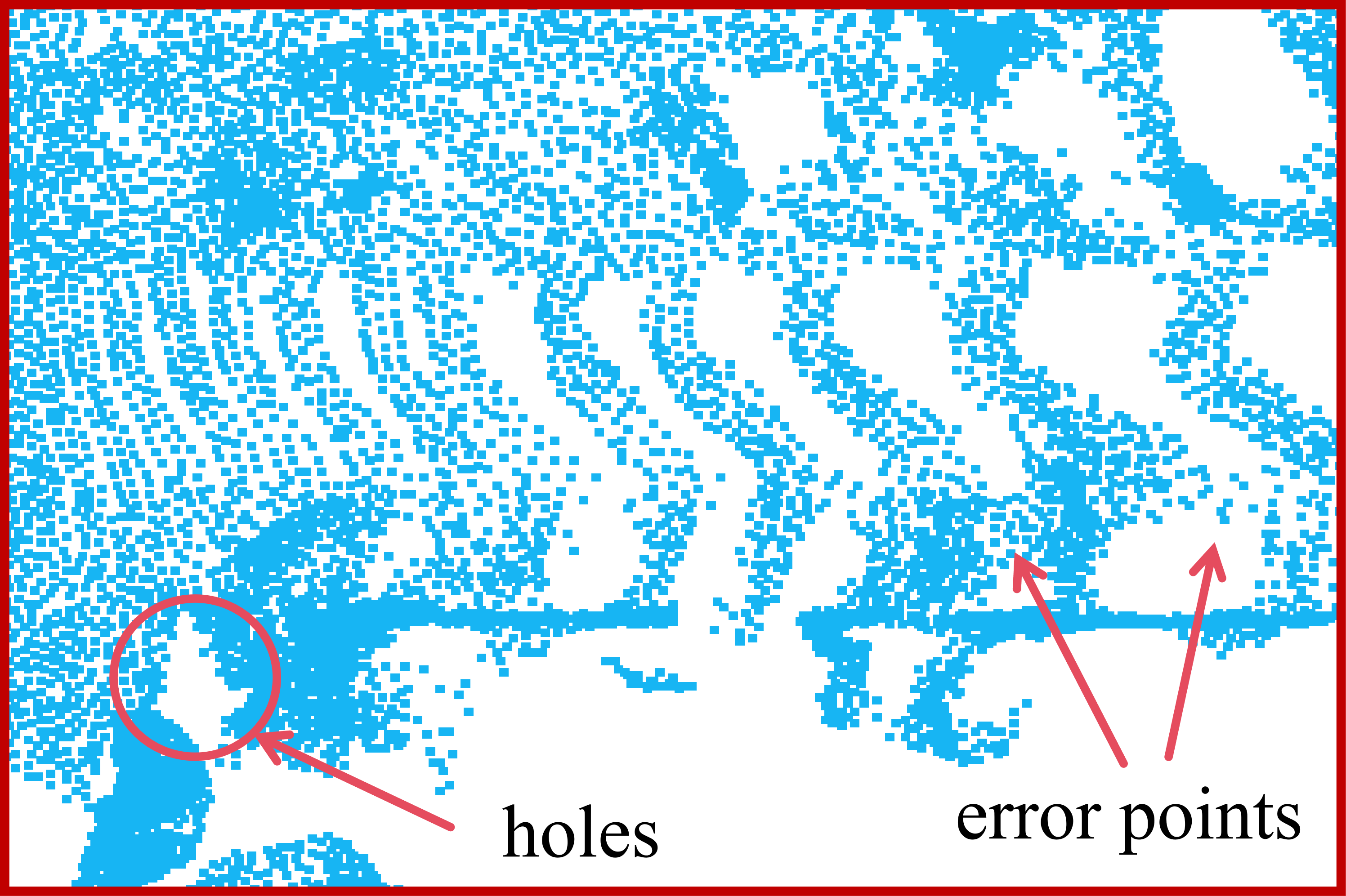}
      \end{tabular}
      \\
      \begin{tabular}{@{}c}
        \rotatebox{90}{(d) PUDM \cite{PUDM}}
      \end{tabular}
      &
      \begin{tabular}{@{}c}
        \includegraphics[width=0.35\textwidth]{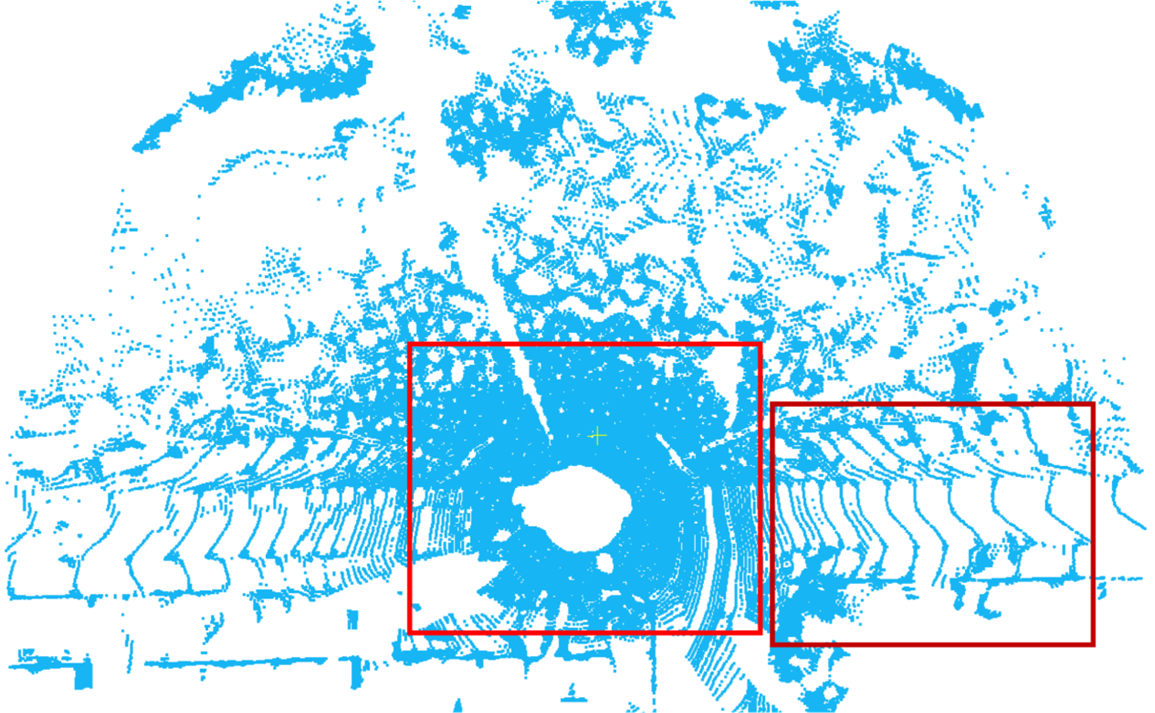}
      \end{tabular}
      &
      \begin{tabular}{@{}c}
        \includegraphics[width=0.25\textwidth]{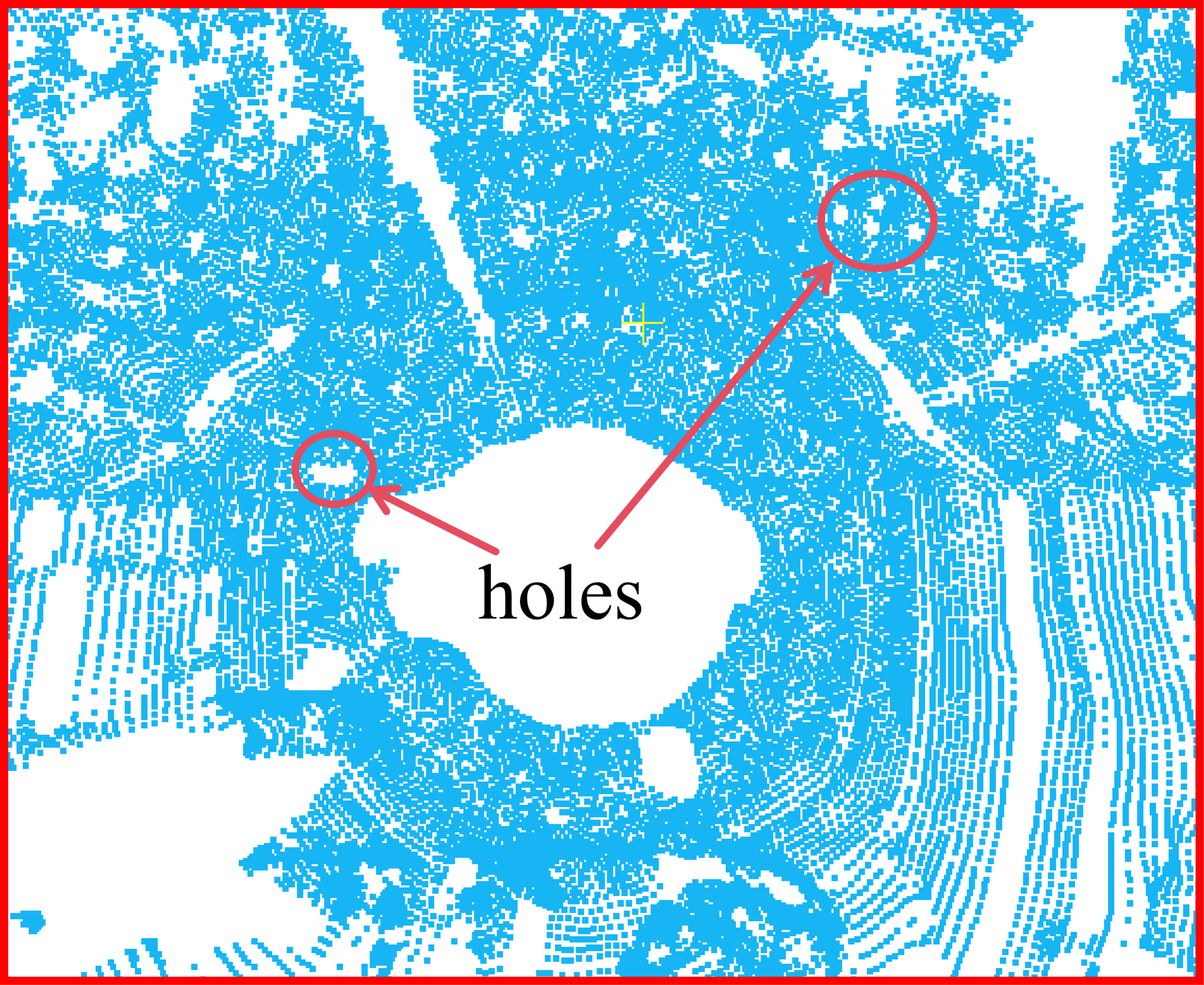}
      \end{tabular}
      &
      \begin{tabular}{@{}c}
        \includegraphics[width=0.3\textwidth]{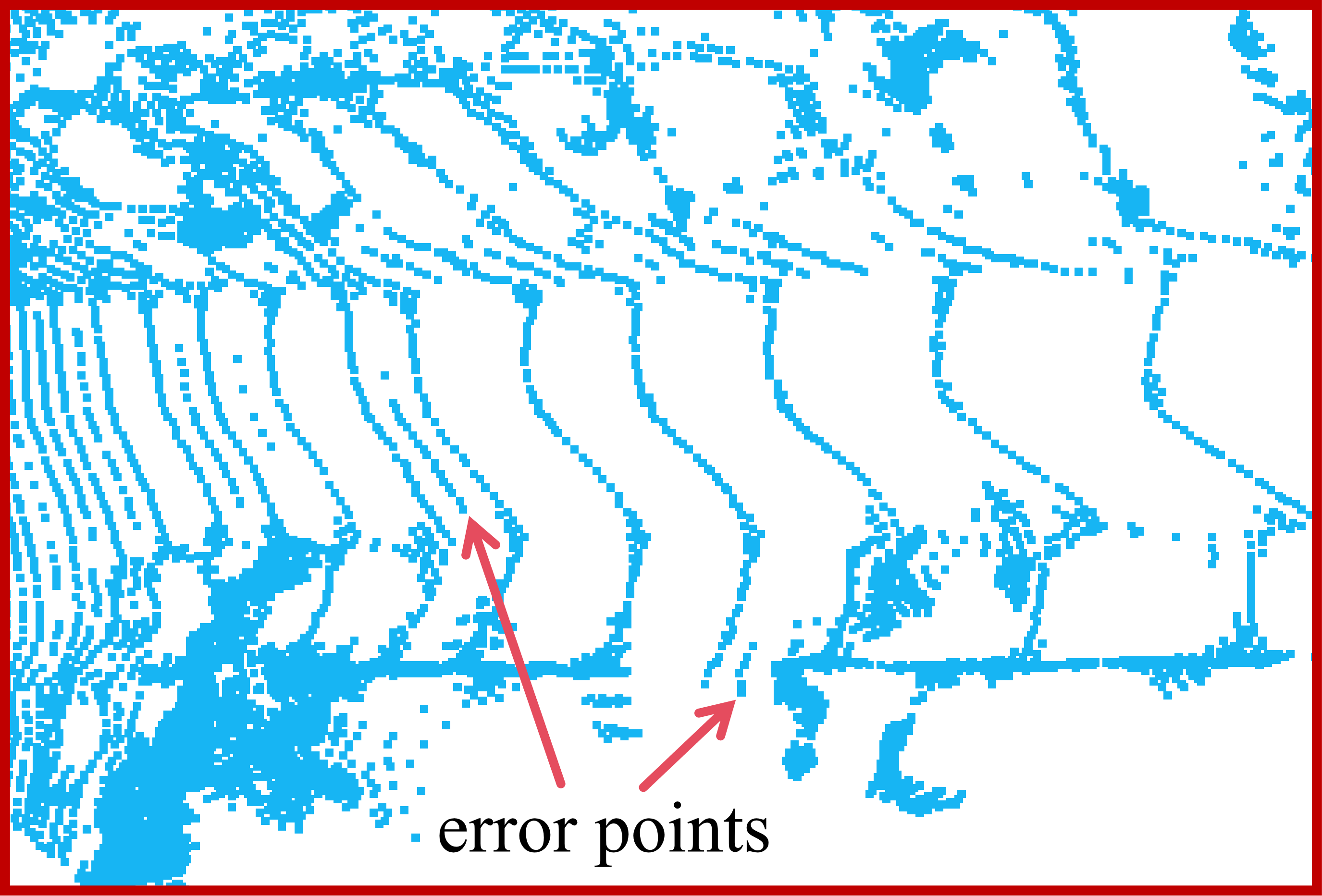}
      \end{tabular}
      \\
      \begin{tabular}{@{}c}
        \rotatebox{90}{(b) TULIP \cite{yang2024tulip}}
      \end{tabular}
      &
      \begin{tabular}{@{}c}
        \includegraphics[width=0.35\textwidth]{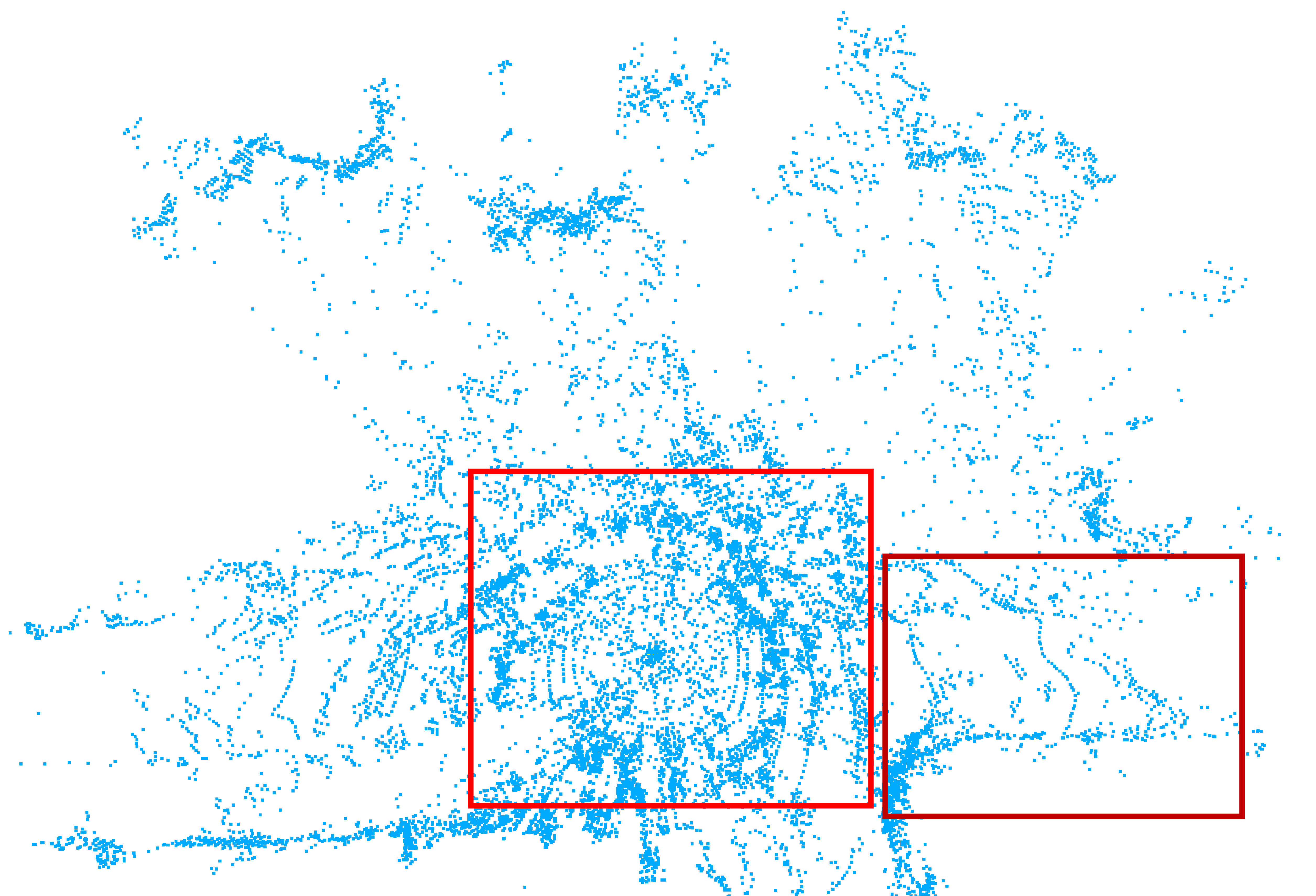}
      \end{tabular}
      &
      \begin{tabular}{@{}c}
        \includegraphics[width=0.25\textwidth]{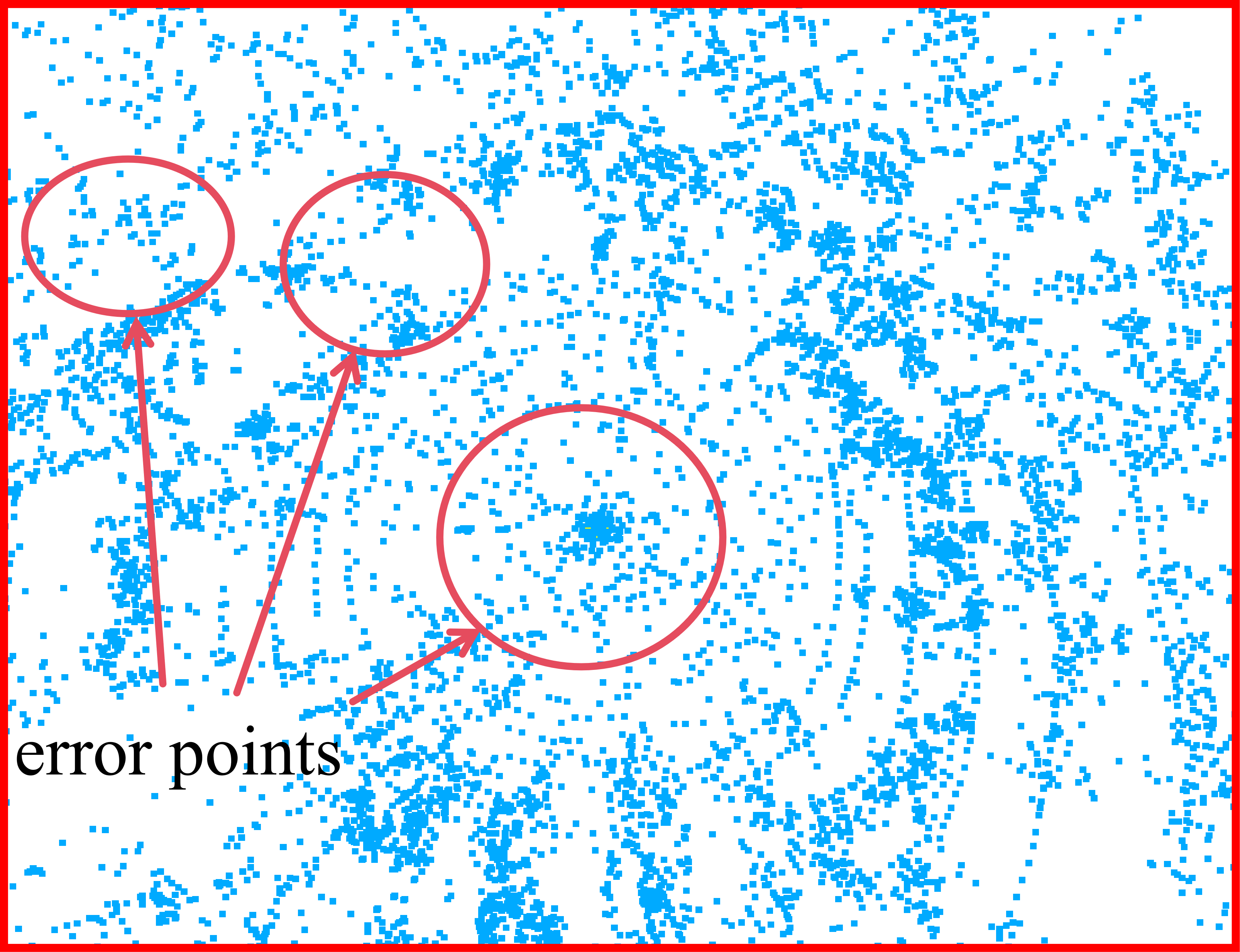}
      \end{tabular}
      &
      \begin{tabular}{@{}c}
        \includegraphics[width=0.3\textwidth]{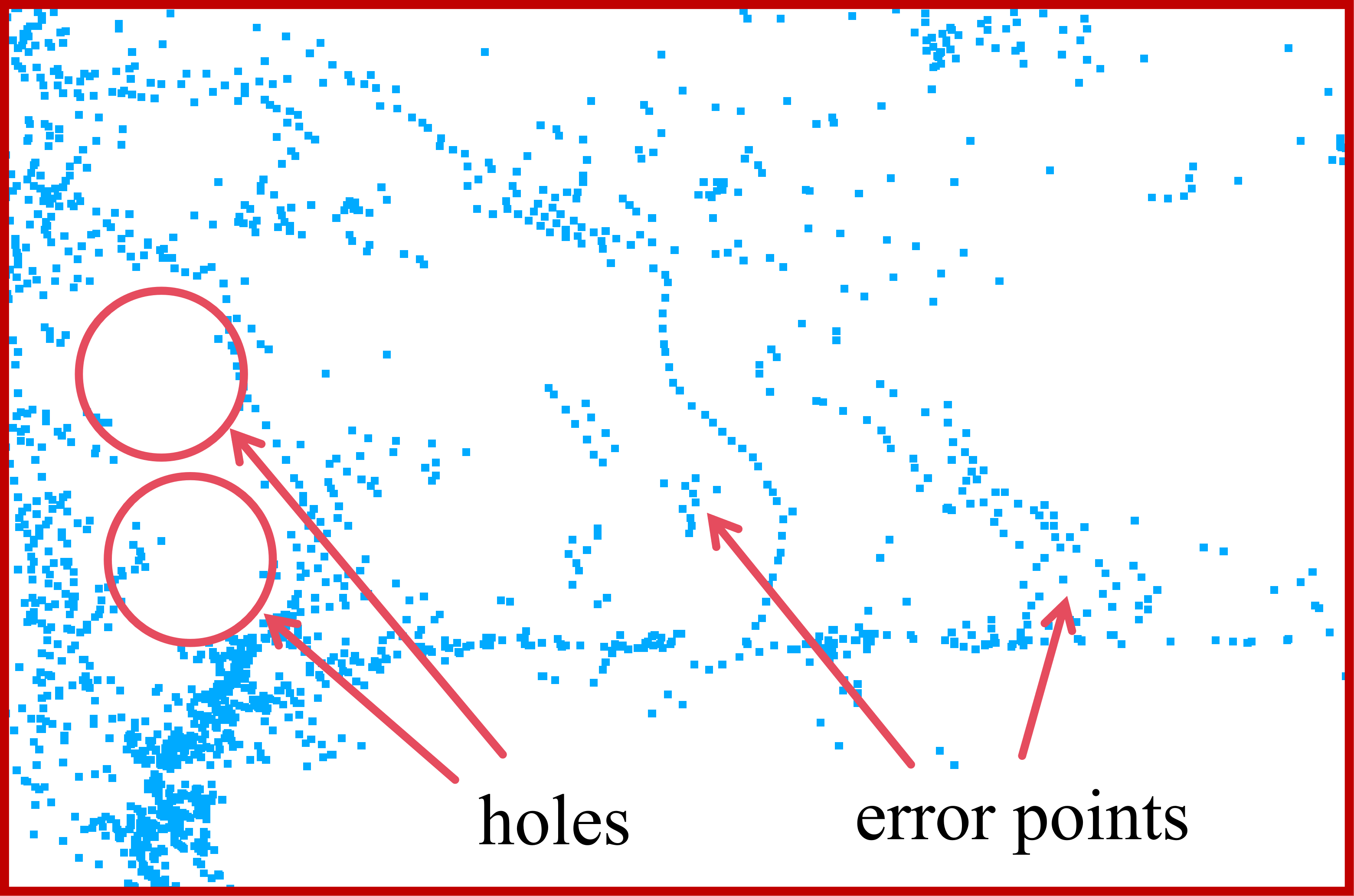}
      \end{tabular}
      \\
      \begin{tabular}{@{}c}
        \rotatebox{90}{(e) \textbf{Ours}}
      \end{tabular}
      &
      \begin{tabular}{@{}c}
        \includegraphics[width=0.35\textwidth]{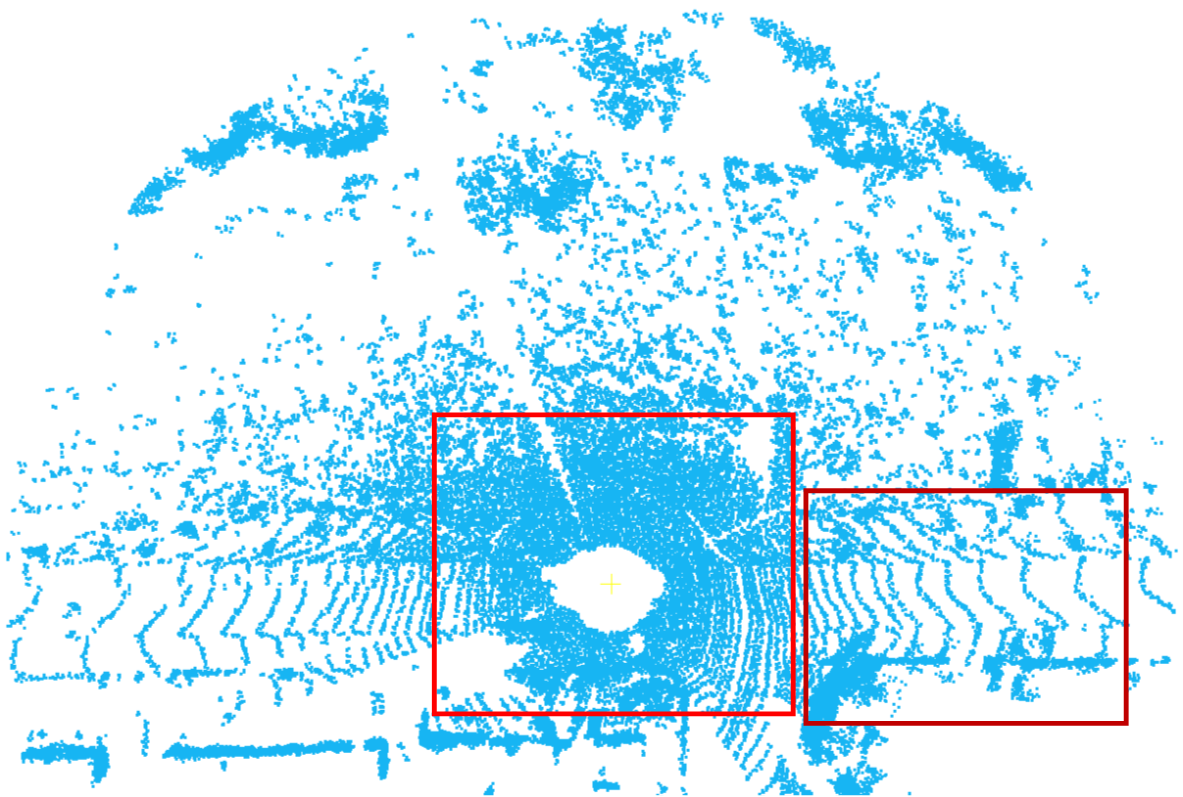}
      \end{tabular}
      &
      \begin{tabular}{@{}c}
        \includegraphics[width=0.25\textwidth]{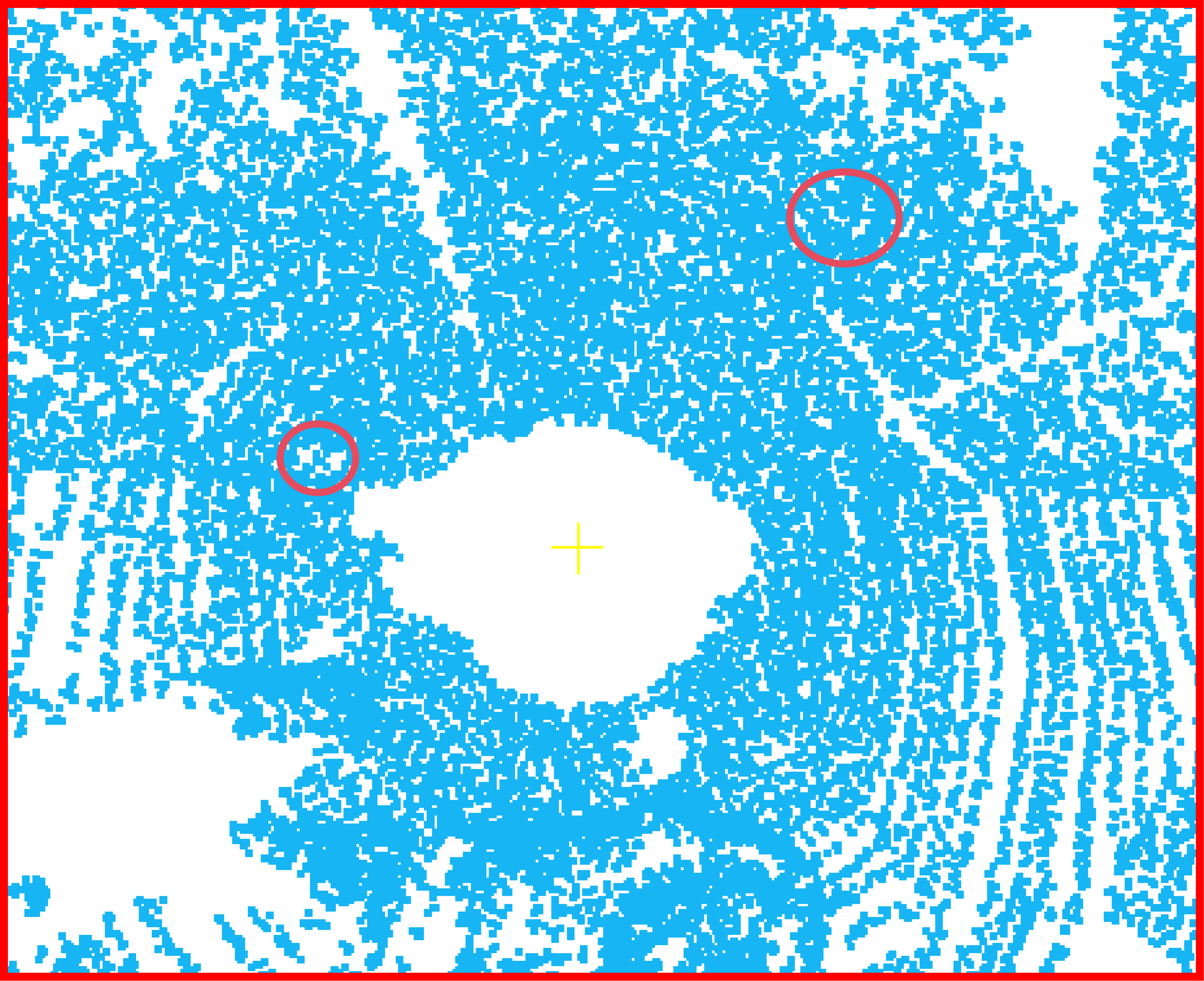}
      \end{tabular}
      &
      \begin{tabular}{@{}c}
        \includegraphics[width=0.3\textwidth]{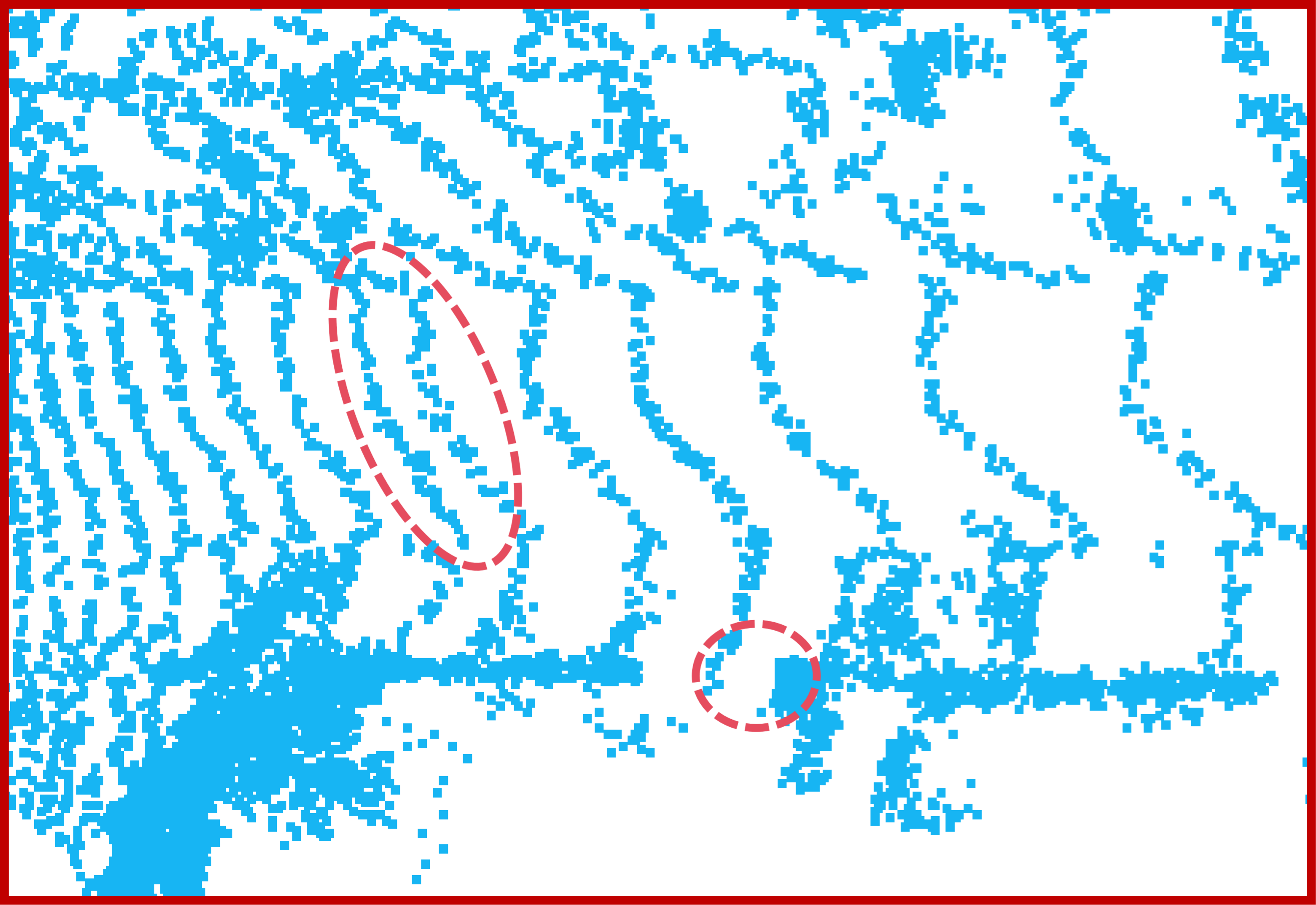}
      \end{tabular}
      \\
  \end{tabular}
  \caption{Qualitative results of our method and state-of-the-art approaches for $4 \times $ upsampling on SemanticKITTI dataset. }
  \label{fig:comp_methods}
\end{figure*}

\renewcommand{\arraystretch}{1.3}
\begin{table*}
  \centering
  \fontsize{9}{9}\selectfont
  \caption{Quantitative comparison between PVNet and state-of-the-art approaches in the supervised setting. $R$ represents the upsampling rate.}
  \label{tab:comparison_benchmarks_supervised}
  \setlength{\tabcolsep}{1.2mm}{
      \begin{tabular}{cccccc@{\hspace{10mm}}cccccccc}
        \toprule
        \multirow{2}{*}{Methods} &\multicolumn{5}{c}{$\times$ 4 (R=4)} &\multicolumn{5}{c}{$\times$ 16 (R=16)} \\
        \cmidrule(r{10mm}){2-6}  \cmidrule(r){7-11}  & CD $\downarrow$ & RCD $\downarrow$ & Recon-RCD $\downarrow$ & Match-RCD $\downarrow$ & Time(s) $\downarrow$ & CD $\downarrow$   &RCD $\downarrow$  &Recon-RCD $\downarrow$  & Match-RCD $\downarrow$ & Time(s) $\downarrow$ \\
        \midrule
        PU-Net \cite{PU_net} & 0.959 & 0.777 & 0.803 & 0.785 & \textbf{0.200} & 1.453 & 1.231 & 1.376 &  1.292 & \textbf{0.060} \\
        PU-GAN \cite{Pu-gan} & 0.663 & 0.593 & 0.613 & 0.602 & 1.460 & 1.021 & 0.650 & 0.659 & 0.670  & 10.53 \\
        Dis-PU \cite{Dis-PU} & 0.277 & 0.421 & 0.440 & 0.426 & 2.201 & 0.342 & 0.585 & 0.627 & 0.606  & 13.43 \\
        PU-GCN \cite{Pu-gcn} & 0.258 & 0.418 & 0.434 & 0.424 & 1.674 & \textbf{0.329} & 0.581 & 0.626 & 0.602  & 11.12 \\
        PUDM \cite{PUDM}     & 616.0 & 16.32 & 17.71 & 16.63 & 5.825 & 616.6 & 16.33 & 17.64 & 16.57  &  5.457 \\
        RepKPU \cite{rong2024repkpu}  & 0.290 &	0.429& 0.440& 0.436& 0.746& 1.067& 0.671& 0.705& 0.682& 0.514 \\
        TULIP \cite{yang2024tulip} & 12.95 & 0.959& 1.162& 1.028& 0.294 & - &- & - &- &-\\
        \textbf{PVNet(Ours)} & \textbf{0.158} & \textbf{0.338}  & \textbf{0.345} & \textbf{0.341}  & 35.88
        & 0.475 & \textbf{0.540} &\textbf{0.567} & \textbf{0.553}  & 37.97\\
        \bottomrule
      \end{tabular}
      }
\end{table*}

\subsection{Comparison Against SOTA Methods}
\label{sec:compare} 
Following previous works \cite{PU_net,Pu-gan,Dis-PU,Pu-gcn}, we first conduct the qualitative comparison based on two fixed upsampling rates, i.e., 4 and 16. The results on SemanticKITTI dataset and KITTI-360 dataset are shown in Table \ref{tab:comparison_benchmarks} and Table \ref{tab:comparison_KITTI-360}, respectively (note that TULIP \cite{yang2024tulip} is architecturally limited to 4$\times$ upsampling and thus excluded from 16$\times$ comparisons). 
Our method significantly outperforms other methods in terms of CD, RCD, Recon-RCD and Match-RCD metrics across two datasets under 4$\times$ and 16$\times$ upsampling. When increasing the upsampling rate from 4$\times$ to 16$\times$, our approach reduces CD values by 0.015 and 0.025 on SemanticKITTI and KITTI-360, respectively, while all baselines exhibit increased CD values. 
Note that PUDM excels at upsampling symmetric objects, but struggles with outdoor point clouds which are typically asymmetric and non-uniform in density. This results in numerous holes and incorrect points deviating from the guiding sparse points, leading to its suboptimal performance across all evaluation metrics.
Regarding the inference time, while our method is noticeably slower than other methods at 4$\times$ upsampling, it becomes competitive at 16$\times$ which surpassing all methods except PU-Net. These results demonstrate that our approach achieves robust performance across both low and high upsampling rates, with particularly pronounced efficiency advantages at higher rates.

Figure  \ref{fig:comp_methods} shows the visual results. We can see that PU-GCN \cite{Pu-gcn} produces more uniformly distributed points but still exhibits numerous holes and error 
points. PUDM \cite{PUDM} generates a significant number of errors far away from 
the guiding sparse points. Additionally, it 
exhibits small holes in the center regions and error points in the distant 
areas. TULIP \cite{yang2024tulip} exhibits some holes and errors in the generated scans and struggles to generate reasonable structures. By contrast, our method achieves superior reconstruction quality by effectively preserving the structure of sparse scans while generating finer details, and our generated point clouds exhibit no holes in the center regions and no error points in the distant areas.

Our method does not require a dense point cloud corresponding to the sparse input as supervision. However, this also means there is no direct ground truth for performance evaluation. To more comprehensively assess our model's capability, we conduct experiments using the original sparse point cloud as ground truth, with input point clouds generated through 4$\times $ and 16$\times $ downsampling of this ground truth. As shown in Table \ref{tab:comparison_benchmarks_supervised}, under this evaluation protocol, our method still demonstrates significant superiority over other approaches at the 4$\times $ upsampling rate. For the 16$\times $ upsampling case, while our model falls behind Dis-PU and PU-GCN in terms of CD, it outperforms all baselines in other quantitative metrics. These results further validate the effectiveness and robustness of our approach.

\renewcommand{\arraystretch}{1.3}
\begin{table*}
  \centering
  \fontsize{9}{9}\selectfont
  \caption{Qualitative results of 4 $\times$ upsampling with different Gaussian noise levels ($\tau$=0.1, 0.2) on our test set. 
  }
  \label{tab:comp_noise_4}
  \setlength{\tabcolsep}{1.2mm}{
      \begin{tabular}{cccccccccccc}
        \toprule
        \multirow{3}{*}{Methods} &\multicolumn{5}{c}{$\tau$ = 0.1 ($\times 4$)} &\multicolumn{2}{c}{$\tau$ = 0.2 ($\times 4$)} \cr
        \cmidrule(lr){2-5}  \cmidrule(lr){6-9}  & CD $\downarrow$ & RCD $\downarrow$ & Recon-RCD $\downarrow$ & Match-RCD $\downarrow$ & CD $\downarrow$   &RCD $\downarrow$  &Recon-RCD $\downarrow$  & Match-RCD $\downarrow$ \cr
        \midrule
        PU-Net \cite{PU_net} & 0.762 & 0.693 & 0.712 & 0.696 & 0.759 & 0.687 & 0.697 & 0.694 \\
        PU-GAN \cite{Pu-gan} & 0.346 & 0.542 & 0.579 & 0.555 & 0.378 & 0.564 & 0.600 & 0.577 \\
        Dis-PU \cite{Dis-PU} & 0.126 & 0.451 & 0.476 & 0.470 & 0.165 & 0.479 & 0.516 & 0.490  \\
        PU-GCN \cite{Pu-gcn} & 0.122 & 0.455 & 0.486 & 0.471 & 0.170 & 0.486 & 0.521 & 0.499 \\
        PUDM \cite{PUDM} & 614.9 & 16.31 & 17.59 & 16.61 & 614.9 & 16.31 & 17.43 & 16.60 \\
        RepKPU \cite{rong2024repkpu}  &\textbf{0.109} & 0.423 & 0.453 & 0.435 & \textbf{0.117} & 0.435 & 0.460 & 0.450 \\
        TULIP \cite{yang2024tulip} & 8.710 & 0.770 & 0.891 & 0.812 & 8.282 & 0.774 & 0.903 & 0.816 \\ 
        \textbf{PVNet(Ours)}  &\underline{0.112} &\textbf{0.354} &\textbf{0.366} &\textbf{0.364} &\underline{0.154} &\textbf{0.384} &\textbf{0.401} &\textbf{0.391} \\
        \bottomrule
      \end{tabular}
      }
\end{table*}

\renewcommand{\arraystretch}{1.3}
\begin{table*}
  \centering
  \fontsize{9}{9}\selectfont
  \caption{Qualitative results of 16 $\times$ upsampling with different Gaussian noise levels ($\tau$=0.1, 0.2) on our test set. 
  }
  \label{tab:comp_noise_16}
  \setlength{\tabcolsep}{1.2mm}{
      \begin{tabular}{cccccccccccc}
        \toprule
        \multirow{3}{*}{Methods} &\multicolumn{5}{c}{$\tau$ = 0.1 ($\times 16$)} &\multicolumn{2}{c}{$\tau$ = 0.2 ($\times 16$)} \cr
        \cmidrule(lr){2-5}  \cmidrule(lr){6-9}  & CD $\downarrow$ & RCD $\downarrow$ & Recon-RCD $\downarrow$ & Match-RCD $\downarrow$ & CD $\downarrow$   &RCD $\downarrow$  &Recon-RCD $\downarrow$  & Match-RCD $\downarrow$ \cr
        \midrule
        PU-Net \cite{PU_net} & 1.805 & 1.013 & 1.046 & 1.033 & 1.766 & 0.997 & 1.044 & 1.010 \\
        PU-GAN \cite{Pu-gan} & 0.658 & 0.713 & 0.776 & 0.733 & 0.687 & 0.730 & 0.773 & 0.758 \\
        Dis-PU \cite{Dis-PU} & 0.154 & 0.711 & 0.768 & 0.736 & 0.184 & 0.723 & 0.770 & 0.749 \\
        PU-GCN \cite{Pu-gcn} & 0.151 & 0.705 & 0.752 & 0.734 & 0.192 & 0.719 & 0.765 & 0.744 \\
        PUDM \cite{PUDM} & 614.8 & 16.31 & 17.57 & 16.61 & 614.8 & 16.31 & 17.28 & 16.76  \\
        RepKPU \cite{rong2024repkpu}  & 0.140 & 0.705 & 0.748 & 0.728 & 0.146 & 0.723 & 0.748 & 0.753 \\
        \textbf{PVNet(Ours)} & \textbf{0.097} & \textbf{0.660} & \textbf{0.711} & \textbf{0.678} & \textbf{0.138} & \textbf{0.674} & \textbf{0.712} & \textbf{0.694} \\
        \bottomrule
      \end{tabular}
      }
\end{table*}

\begin{figure*}
  \centering
    \begin{tabular}{@{}c@{}c@{}c@{}c}
      \begin{tabular}{@{}c}
        \includegraphics[width=0.24\linewidth]{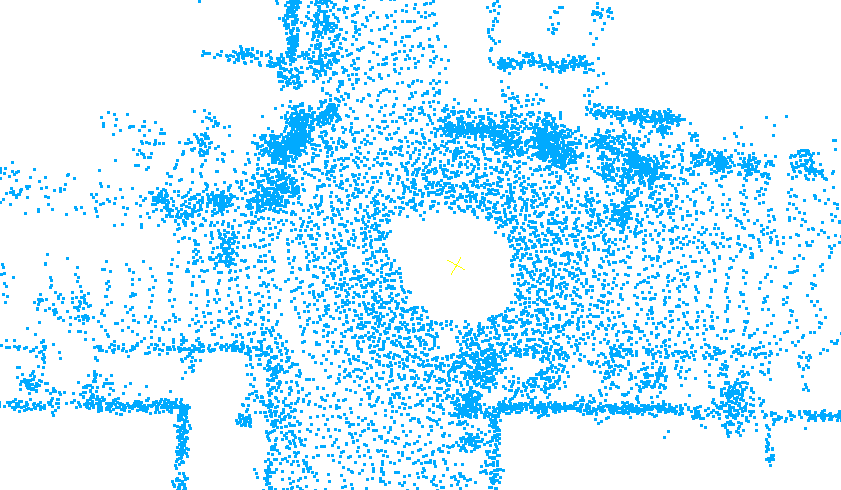}
      \end{tabular}
      &
      \begin{tabular}{@{}c}
        \includegraphics[width=0.24\linewidth]{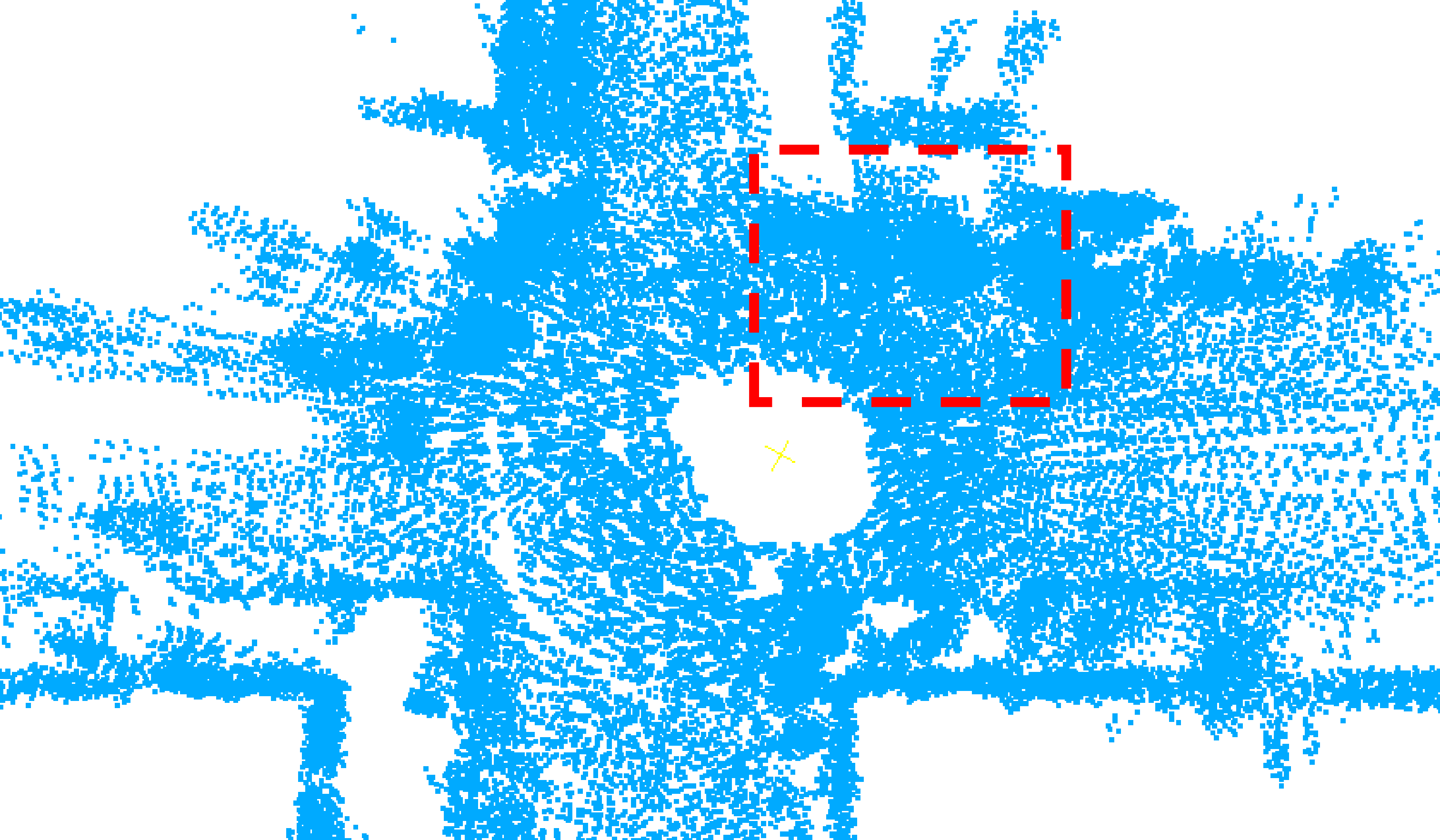}
      \end{tabular}
      &
      \begin{tabular}{@{}c}
        \includegraphics[width=0.24\linewidth]{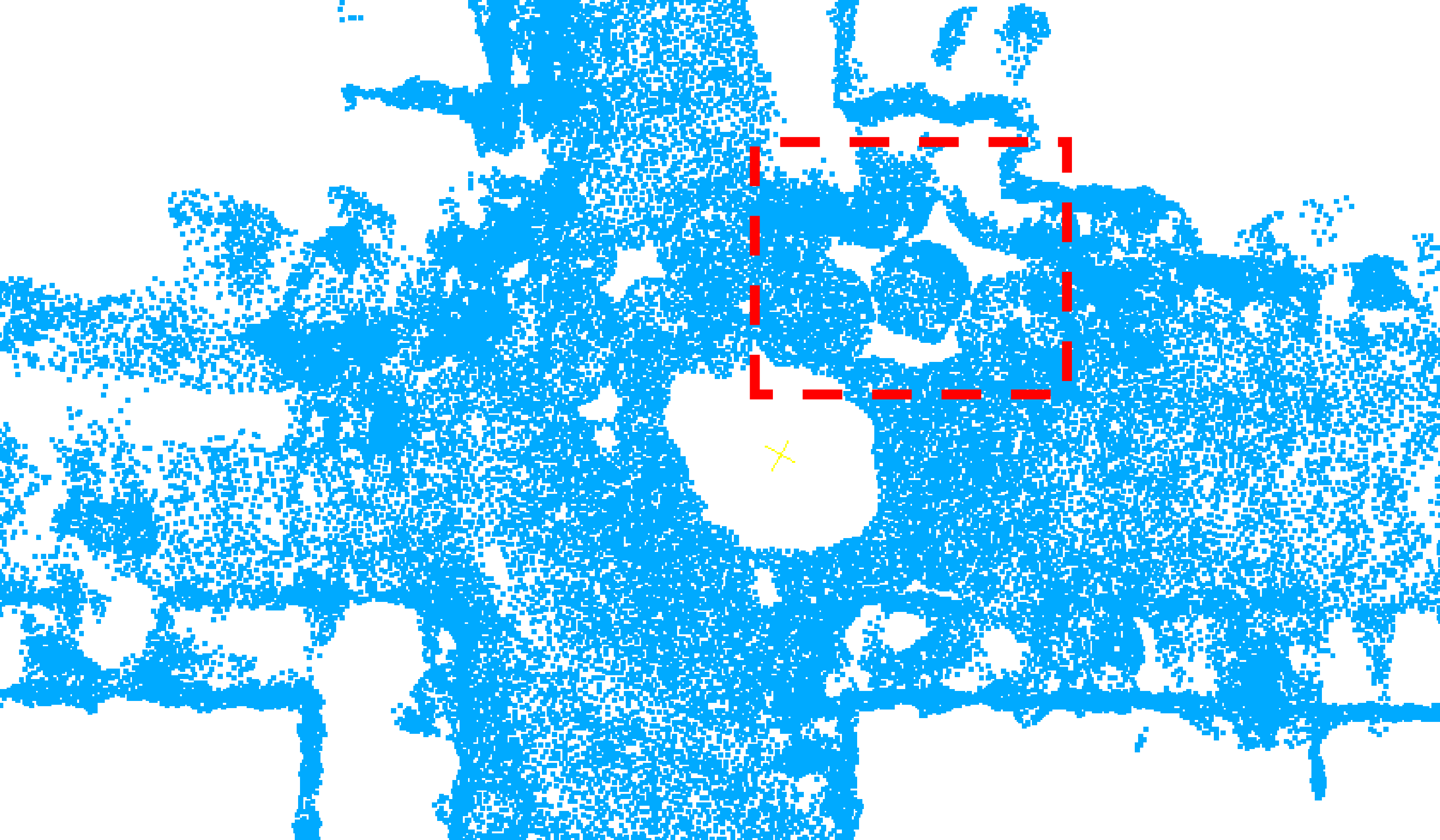}
      \end{tabular}
      &
      \begin{tabular}{@{}c}
        \includegraphics[width=0.24\linewidth]{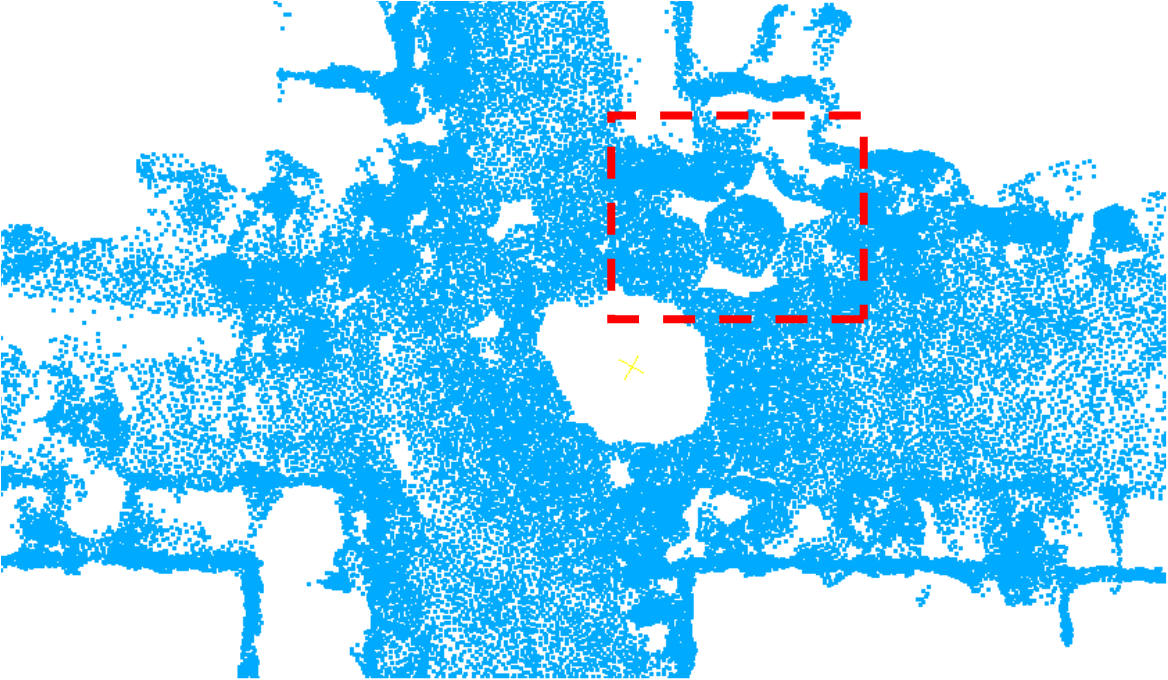}
      \end{tabular}
      \\
      (a) Input
      &
      (b) PU-Net \cite{PU_net}
      & 
      (c) Dis-PU \cite{Dis-PU}
      &
      (d) PU-GCN \cite{Pu-gcn}
      \\
      \begin{tabular}{@{}c}
        \includegraphics[width=0.24\linewidth]{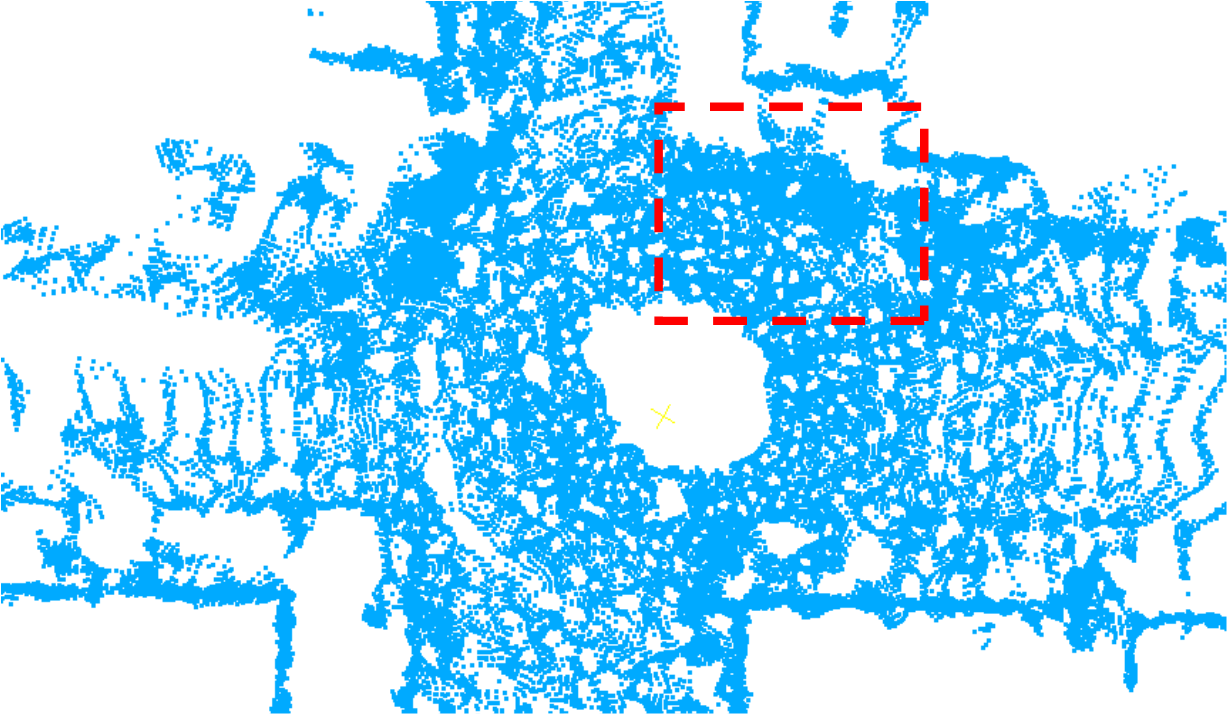}
      \end{tabular}
      &
      \begin{tabular}{@{}c}
        \includegraphics[width=0.242\linewidth]{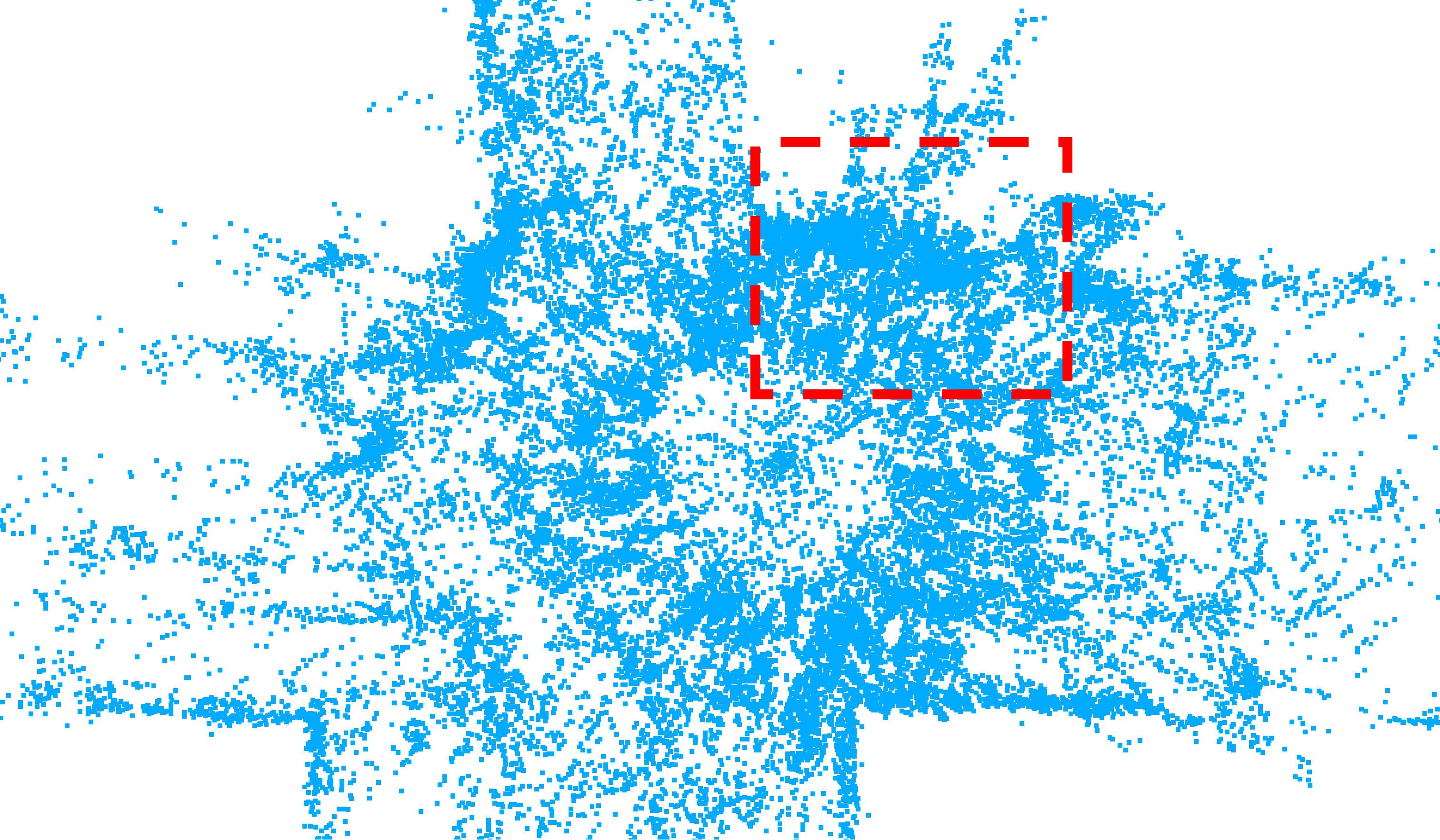}
      \end{tabular}
      &
      \begin{tabular}{@{}c}
        \includegraphics[width=0.24\linewidth]{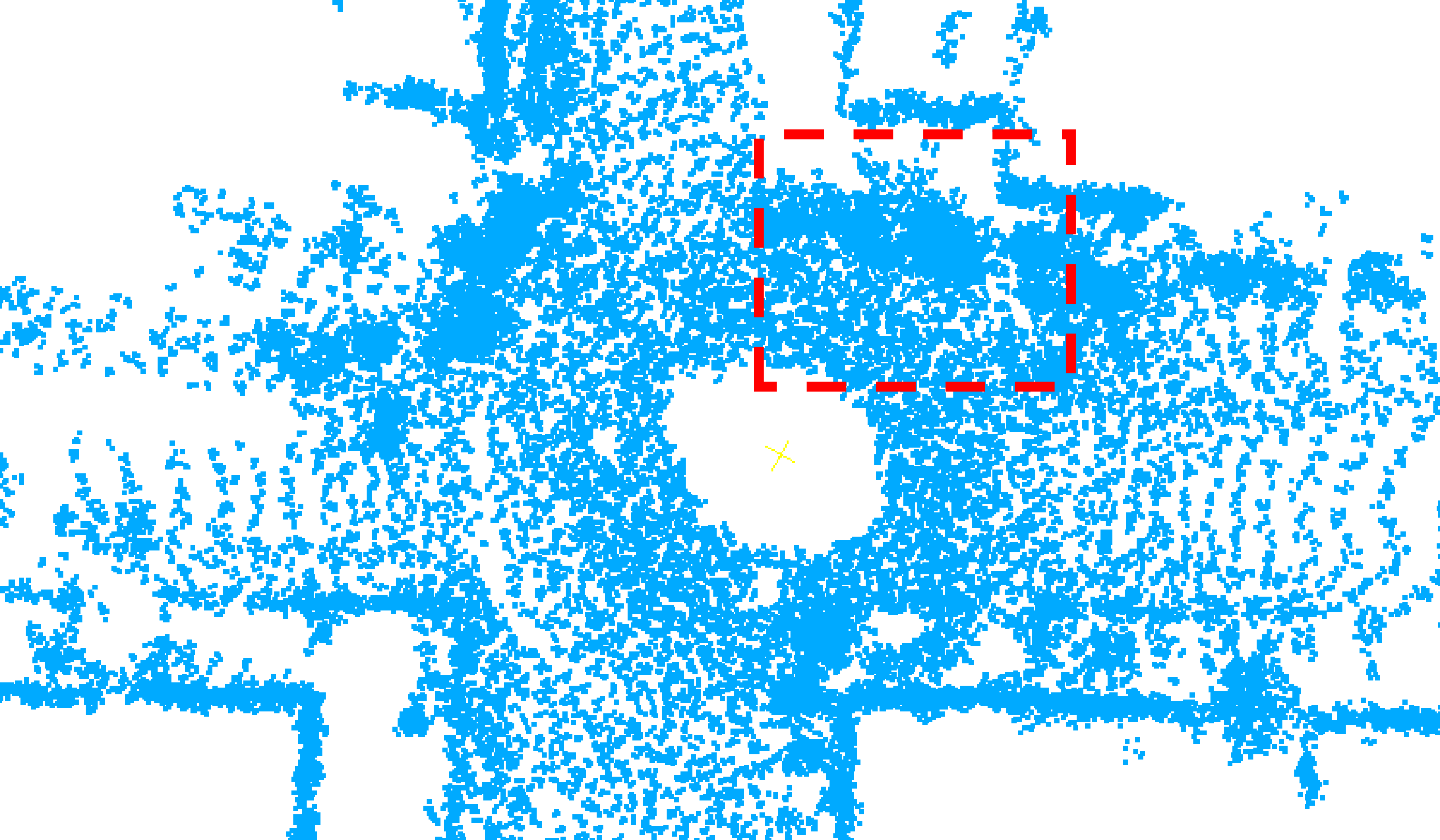}
      \end{tabular}
      &
      \begin{tabular}{@{}c}
        \includegraphics[width=0.24\linewidth]{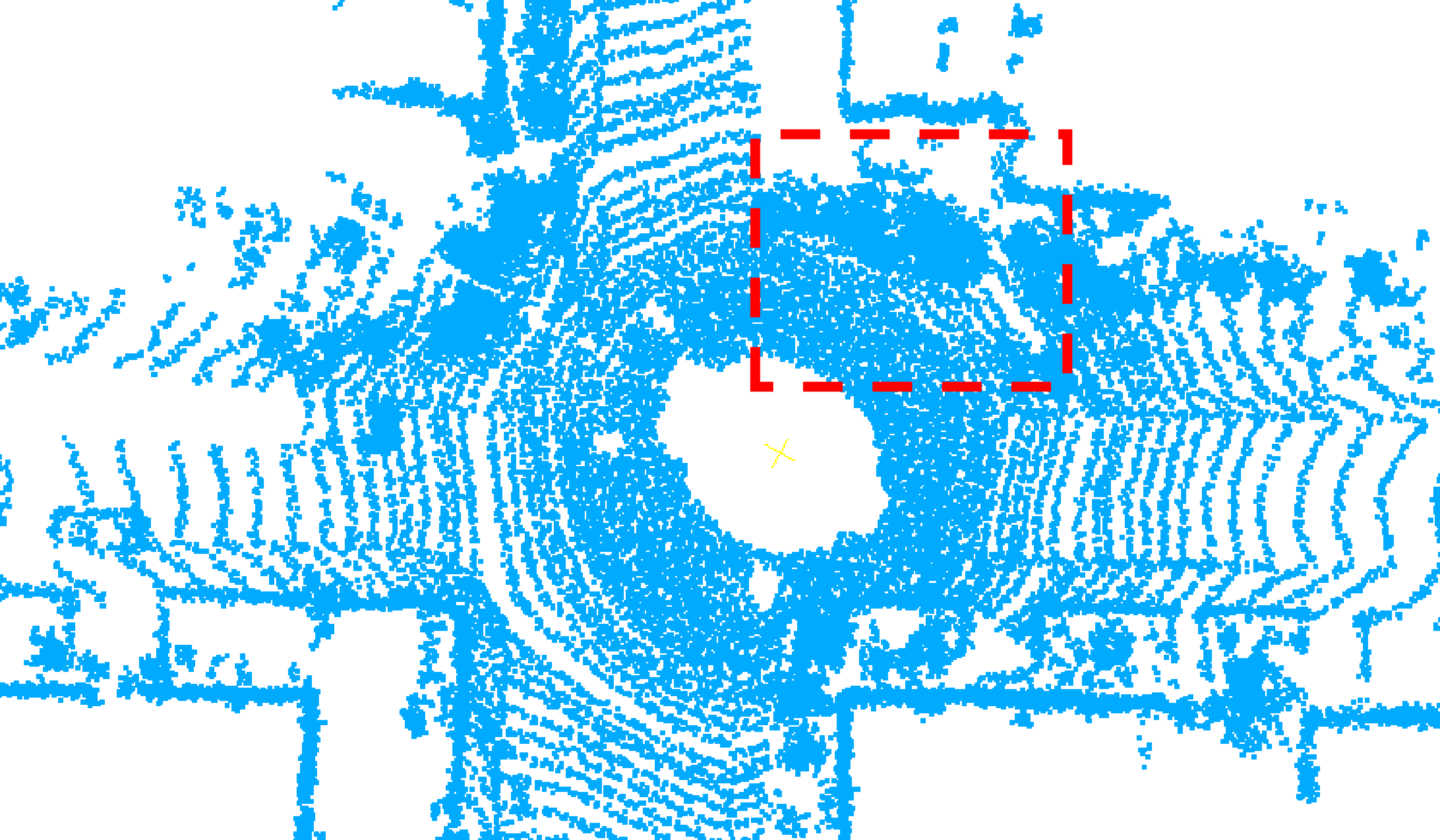}
      \end{tabular}
      \\
      (e) PUDM \cite{PUDM}
      &
      (f) TULIP \cite{yang2024tulip}
      &
      (g) Ours\_noise
      & 
      (h) Ours\_std
  \end{tabular}
  \caption{Comparisons of $4 \times$  upsampling on our test set under Gaussian noise level $\tau$ = 0.2. 
  Our\_noise represents our upsampled results with Gaussian noise, 
  Our\_std represents our upsampled results without adding Gaussian noise.}
  \label{fig:comp_noise}
\end{figure*}

\subsection{Results of Robustness Test}
\label{sec:robust_test}
\textbf{Gaussian Noise.} To demonstrate the robustness of our method, we 
further evaluate the performance under the perturbation of varying noise levels. 
Specifically, we generate sparse point clouds with Gaussian noise sampled from 
a standard Gaussian distribution $\mathcal N(0,1)$ scaled by different noise 
levels $\tau$. 

We conduct the quantitative comparisons under $4\times$ and $16\times$ upsampling as 
shown in Table \ref{tab:comp_noise_4}, \ref{tab:comp_noise_16}, respectively. 
The results show that our method still achieves the best performance across 
multiple noise perturbations ($\tau$=0.1, 0.2), which further demonstrates the 
effectiveness of our method. The visual results are shown in Figure \ref{fig:comp_noise}, 
which demonstrates that our method effectively achieves high-quality upsampled 
results with fine-grained details and clearer structures when processing inputs 
with varying levels of noise.

\renewcommand{\arraystretch}{1.3}
\begin{table*}
    \centering
    \fontsize{9}{9}\selectfont
    \caption{Quantitative comparison between our method and PUDM \cite{PUDM} with arbitrary upsampling rates.}
    \label{tab:comp_arbitrary_rate}
    \setlength{\tabcolsep}{1.2mm}{
      \begin{tabular}{cccccccccc}
        \toprule
        \multirow{3}{*}{Methods} &\multicolumn{4}{c}{PUDM \cite{PUDM}} &\multicolumn{4}{c}{\textbf{PVNet(Ours)}}	  \cr
        \cmidrule(lr){2-5}   \cmidrule(lr){6-9} 
        & CD $\downarrow$ & RCD $\downarrow$ & R-RCD $\downarrow$ & M-RCD $\downarrow$ & CD $\downarrow$ & RCD $\downarrow$ & R-RCD $\downarrow$ & M-RCD $\downarrow$  \cr
        &$10^{-3}$ &$10^{-3}$ &$10^{-3}$ &$10^{-3}$ &$10^{-3}$ &$10^{-3}$ &$10^{-3}$ &$10^{-3}$ \cr
        \midrule
        $\times$ 5  & 614.432  & 16.314 & 17.462  & 16.600 & \textbf{0.093} & \textbf{0.395} & \textbf{0.415} & \textbf{0.407} \\
        $\times$ 6  & 614.494 & 16.314 & 17.423 & 16.654 & \textbf{0.091} & \textbf{0.439} & \textbf{0.463} &\textbf{0.449}  \\
        $\times$ 7  & 614.499 & 16.314 & 17.662 & 16.568 & \textbf{0.089} & \textbf{0.475} & \textbf{0.510} & \textbf{0.485} \\
        $\times$ 8  & 614.507 & 16.314 & 17.670 & 16.548 & \textbf{0.088} & \textbf{0.509} & \textbf{0.540} & \textbf{0.526} \\
        $\times$ 10 & 614.523 & 16.314 & 17.568 & 16.639 & \textbf{0.088} & \textbf{0.509} & \textbf{0.545} & \textbf{0.524} \\
        \bottomrule
      \end{tabular}
      }
\end{table*}

\begin{figure*}
  \centering
  \includegraphics[width=125mm, height=110mm]{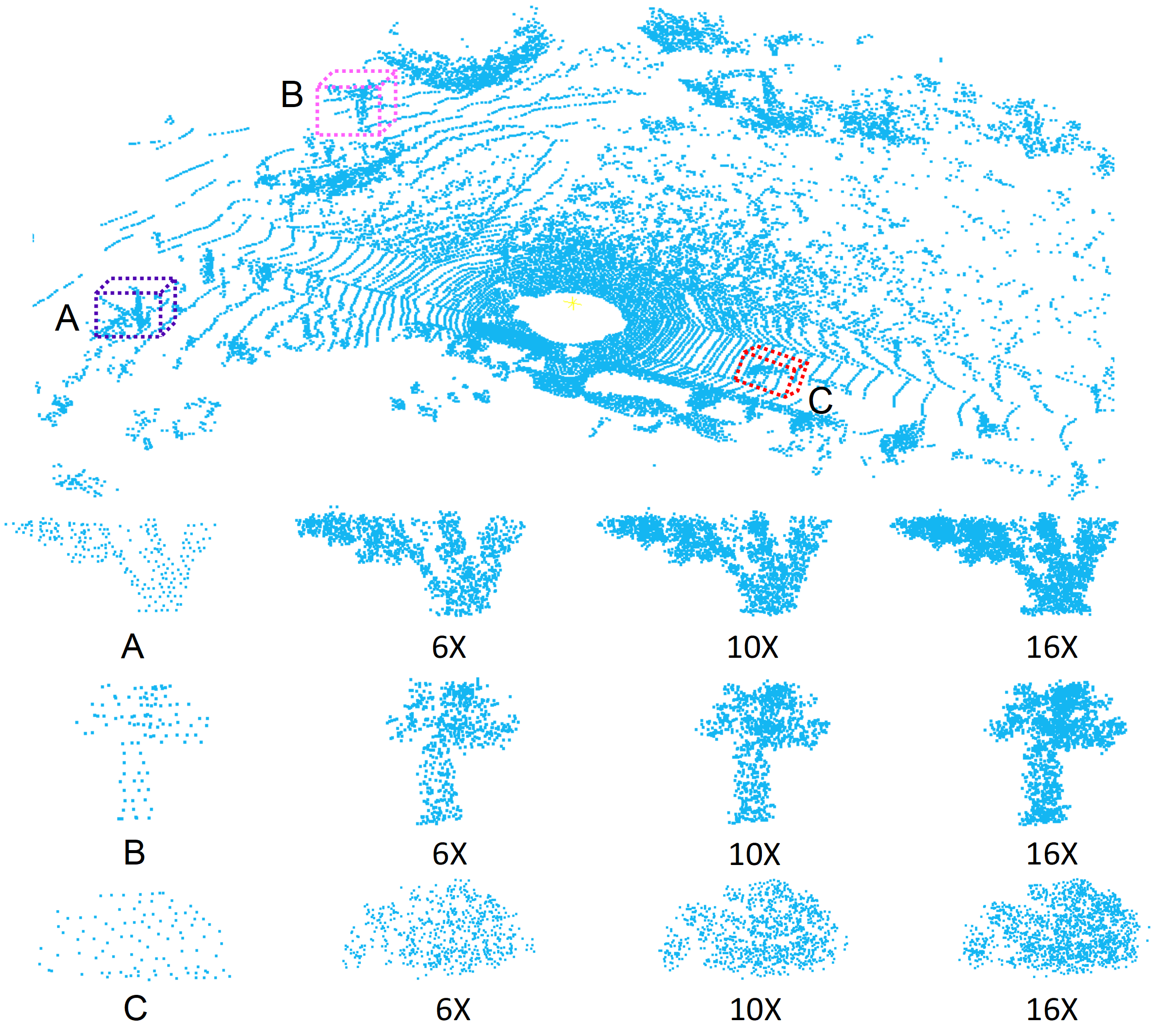}
  \caption{Visual results of our method across multiple upsampling rates. }
  \label{fig:comp_arbitrary_rate}
\end{figure*}
\textbf{Arbitrary Upsampling Rates.} Our method is capable of generating 
high-quality upsampled results based on arbitrary upsampling rates. 
PUDM \cite{PUDM} can also achieve arbitrary upsampling with impressive results. 
Therefore, to further validate the robustness of our method, we conduct the 
comparison with PUDM \cite{PUDM} under arbitrary upsampling rates as shown in 
Table \ref{tab:comp_arbitrary_rate}. From the quantitative results, we can see that 
our method surpasses PUDM by a large margin across multiple upsampling scales. 

Figure \ref{fig:comp_arbitrary_rate} illustrates the detailed results of our arbitrary-scale point cloud upsampling. Our method generates high-quality 
point clouds with denser, more uniform distributions, finer details, and clearer 
structures across multiple upsampling rates.   

\renewcommand{\arraystretch}{1.5}
\begin{table}
    \centering
    \fontsize{9}{9}\selectfont
    \caption{Quantitative comparison between our method and ScalingDiff \cite{ScaleDiff} in terms of object completion in outdoor scenes.}
    \label{tab:comp_scene}
    \setlength{\tabcolsep}{1.2mm}{
      \begin{tabular}{cccc}
        \toprule
         Method  & CD [m] $\downarrow$ & F-Score $\uparrow$  & Params (MB) \cr
        \midrule
         ScalingDiff \cite{ScaleDiff}  & 1.173    & 6.529   & 130.690 \\
         \textbf{Ours} & \textbf{0.551}  & \textbf{11.945}  & \textbf{42.516} \\
        \bottomrule
      \end{tabular}
      }
\end{table}

\begin{figure}
  \centering
    \begin{tabular}{@{}c@{}c@{}c@{}c}
      \begin{tabular}{@{}c}
        \includegraphics[width=19mm, height=20mm]{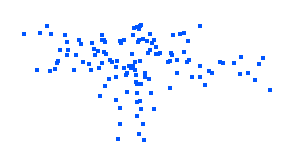}
      \end{tabular}
      &
      \begin{tabular}{@{}c}
        \includegraphics[width=19mm, height=20mm]{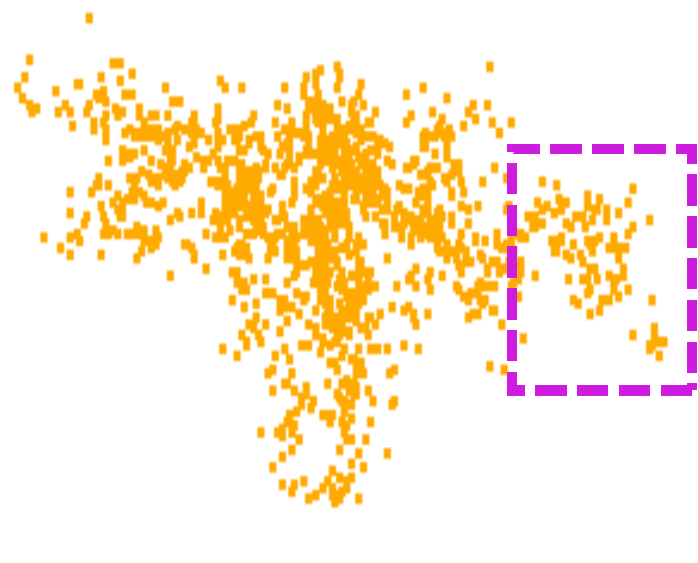}
      \end{tabular}
      &
      \begin{tabular}{@{}c}
        \includegraphics[width=19mm, height=20mm]{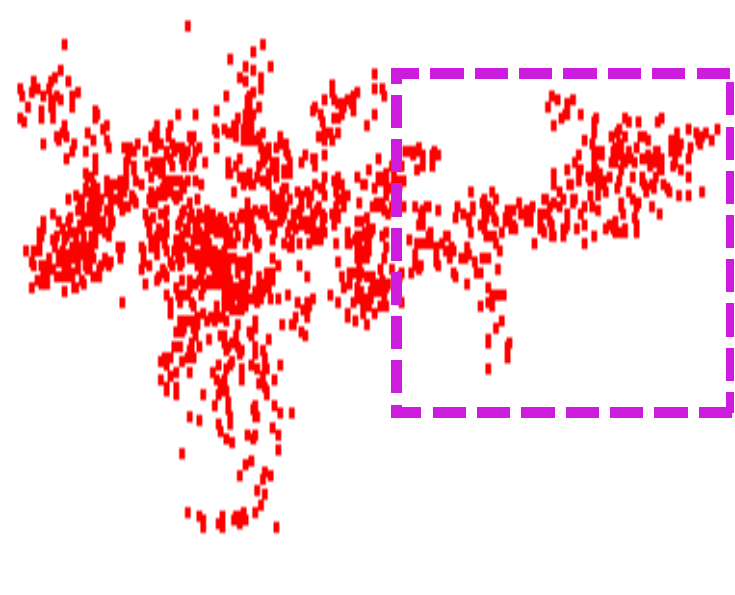}
      \end{tabular}
      &
      \begin{tabular}{@{}c}
        \includegraphics[width=19mm, height=20mm]{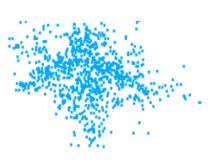}
      \end{tabular}
      \\
      \begin{tabular}{@{}c}
        \includegraphics[width=19mm, height=13.5mm]{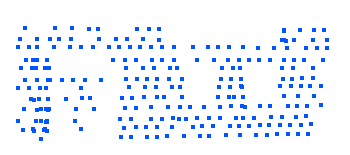}
      \end{tabular}
      &
      \begin{tabular}{@{}c}
        \includegraphics[width=19mm, height=13.5mm]{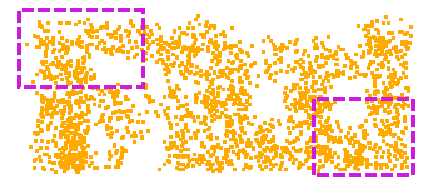}
      \end{tabular}
      &
      \begin{tabular}{@{}c}
        \includegraphics[width=19mm, height=13.5mm]{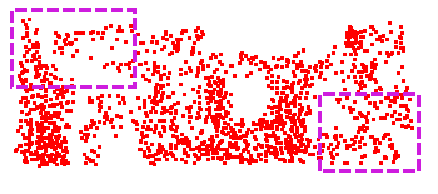}
      \end{tabular}
      &
      \begin{tabular}{@{}c}
        \includegraphics[width=19mm, height=13.5mm]{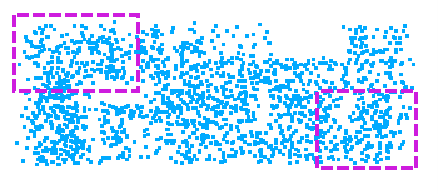}
      \end{tabular}
      \\
      \begin{tabular}{@{}c}
        \includegraphics[width=19mm, height=11.5mm]{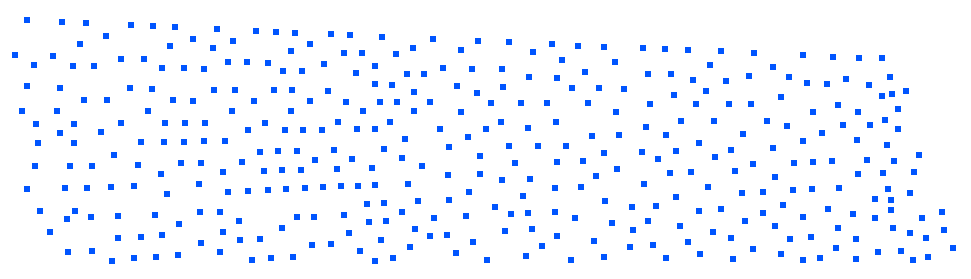}
      \end{tabular}
      &
      \begin{tabular}{@{}c}
        \includegraphics[width=19mm, height=11.5mm]{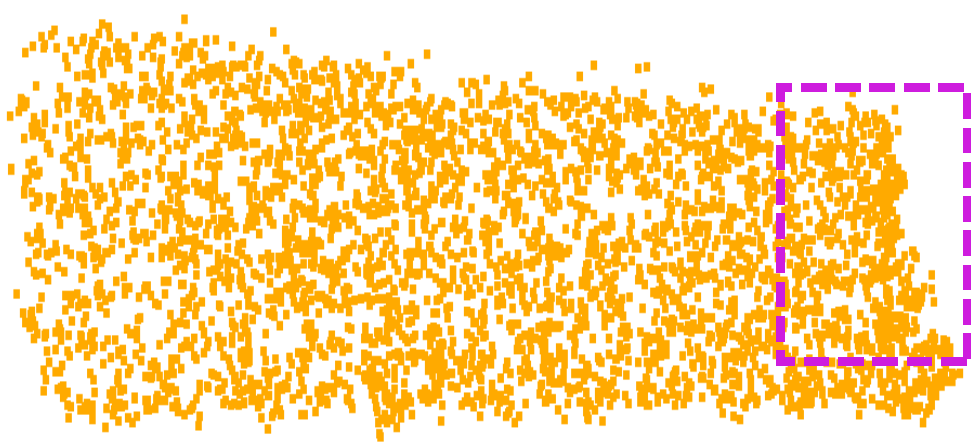}
      \end{tabular}
      &
      \begin{tabular}{@{}c}
        \includegraphics[width=19mm, height=11.5mm]{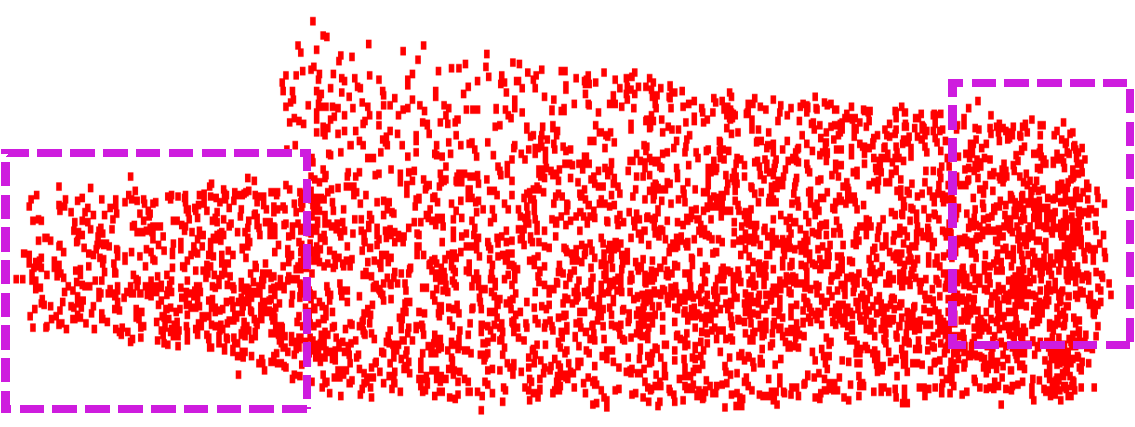}
      \end{tabular}
      &
      \begin{tabular}{@{}c}
        \includegraphics[width=19mm, height=11.5mm]{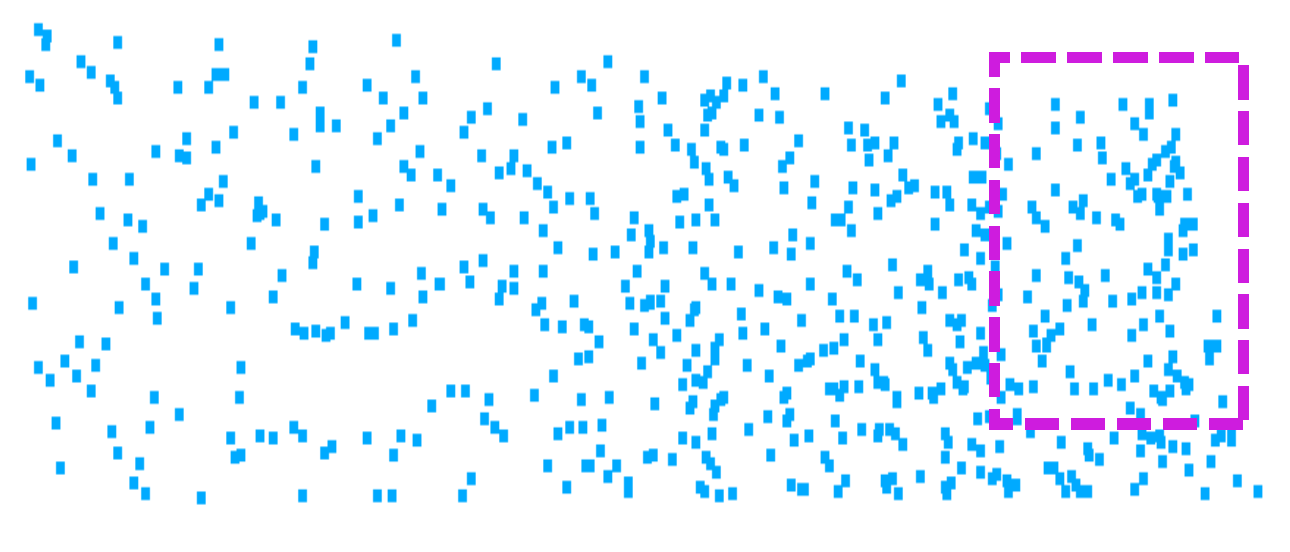}
      \end{tabular}
      \\
      (a) Input
      &
      (b) Ours
      &
      (c) ScalingDiff \cite{ScaleDiff}
      & 
      (d) GT
  \end{tabular}
  \caption{Object completion results of our method and ScalingDiff \cite{ScaleDiff}. GT represents ground truth.}
  \label{fig:comp_recon}
\end{figure}

\subsection{Object Completion}
\label{sec:object_comp}
ScalingDiff \cite{ScaleDiff} is the first method to adopt the diffusion model-based framework for achieving outdoor scene completion and demonstrates its completion performance on SemanticKITTI \cite{SemanticKITTI1}. Our method can not only generate upsampled results with arbitrary upsampling rates but also complete objects in scenes. To explore the completion performance of our method, we conduct detailed comparisons with ScalingDiff \cite{ScaleDiff} and employ the F-Score metric to measure the quality of completed objects on the validation set. For fair comparisons, we adopt the same experimental settings as ScalingDiff. Concretely, the point numbers of the incomplete input and the complete output are 18000 and 180000, respectively.  The point number of the input synthesized point cloud in our framework is also 180000. 
Note that ScalingDiff simultaneously completes both the ground and objects in a scene. However, most existing 3D perception tasks primarily focus on objects located on the ground, such as 3D object detection and instance segmentation. Therefore, we only evaluate the model's ability to complete objects above the ground. As shown in Table \ref{tab:comp_scene}, our method significantly outperforms ScalingDiff in completing objects.

The visual results of completed objects are shown in Figure \ref{fig:comp_recon}. ScalingDiff \cite{ScaleDiff} often generates more unreasonable details and errors. In contrast, our method generates high-quality objects with fine-grained details, as well as more complete and reasonable structures. 

\subsection{Ablation Studies}
\label{sec:ablation}
In this section, we conduct extensive experiments to comprehensively validate the effectiveness of the main components of our method and analyze the sensitivities of key parameters. We use two widely used metrics (CD and F-Score) to measure the results on the validation set. 

\renewcommand{\arraystretch}{1.5}
\begin{table}
  \centering
  \fontsize{8.5}{8.5}  \selectfont
  \caption{Effectiveness of the voxel completion module and the U-Net module.}
    \setlength{\tabcolsep}{1.2mm}{
      \begin{tabular}{cc|ccc}
        \toprule
        Voxel Completion  & U-Net  & CD [m] $\downarrow$  & F-Score $\uparrow$  & Param (MB) \cr
        \midrule
        $\times$      & $\times$       & 0.436   & 6.504  & 5.194  \\
        $\checkmark$  & $\times$       & 0.687   & 7.347  & 19.736 \\
        $\checkmark$  & $\checkmark$   & \textbf{0.425} & \textbf{8.021}  & 42.516 \\
        \bottomrule
      \end{tabular}
      }
     \label{tab:comp_modules}
\end{table}

\renewcommand{\arraystretch}{1.5}
\begin{table}
	  \centering
	  \fontsize{9}{9}\selectfont
	  \caption{Effectiveness of the Point-Voxel Interaction module.}
	  \label{tab:comparison_pvi}
    \setlength{\tabcolsep}{2mm}{
      \begin{tabular}{ccc}
        \toprule
        Component & CD [m] $\downarrow$  & F-Score $\uparrow$ \\
        \midrule
        w/o Point-Voxel Interaction                     & 0.836  & 1.966   \\
        w/ Point-Voxel Interaction    & \textbf{0.425}  & \textbf{8.021}  \\
        \bottomrule
      \end{tabular}
      }
\end{table}

\textbf{Key Components.}
We first analyze the effectiveness of the voxel completion module and the U-Net module in our framework. As shown in Table \ref{tab:comp_modules}, without both of them, we get worse results. After incorporating the voxel completion module, we observe an improvement in the quality of the generated results as indicated by the F-Score metric, although the CD metric shows a decline. By incorporating both modules, we achieve the best results across all metrics. This demonstrates that both the voxel completion module and the U-Net module contribute positively to our framework. 

Next, we conduct an ablation study on the point-voxel interaction module. As shown in Table \ref{tab:comparison_pvi}, incorporating this module improves the CD and F1-score by 0.411 and 6.055, respectively. Such a significant improvement demonstrates the critical role of point-voxel interaction in our framework. 

\renewcommand{\arraystretch}{1.5}
\begin{table}
	  \centering
	  \fontsize{9}{9}\selectfont
	  \caption{Impact of the normalization layers and bias.}
	  \label{tab:comparison_BN_Bias}
    \setlength{\tabcolsep}{2mm}{
      \begin{tabular}{ccc}
        \toprule
        Component & CD [m] $\downarrow$  & F-Score $\uparrow$ \\
        \midrule
        w/ Bias                     & 0.649  & 7.592   \\
        w/ Normalization            & 0.439  & 7.061   \\
        w/ Normalization + Bias     & 0.652  & 7.909   \\
        w/o Normalization + Bias    & \textbf{0.425}  & \textbf{8.021}  \\
        \bottomrule
      \end{tabular}
      }
\end{table}
\textbf{Normalization Layers and Bias.}
In our voxel completion module, we discard the normalization layers and bias to preserve the spatial relation between voxels and the structural integrity within voxels during optimization. In this section, we conduct ablation experiments to validate the impact of the normalization layers and/or bias on the final results. As shown in Table \ref{tab:comparison_BN_Bias}, after adding normalization layers and bias, the model’s performance on CD and F-Score decreased by 0.227 and 0.112, respectively. Moreover, adding bias or normalization layers both degrade the performance of the proposed method. 
These findings indicate that the normalization layer and bias have a negative impact on the quality of upsampled results. 
Although removing these operations from the model is not innovative, our work is the first to identify and analyze the negative impact of normalization and bias on the 3D LiDAR point cloud upsampling task, which provides valuable insights for future 3D scene analysis and completion tasks.

\renewcommand{\arraystretch}{1.5}
\begin{table}
    \centering
    \fontsize{9}{9} \selectfont
    \caption{Analysis of the number of neighboring voxels in the point-voxel interaction module. $K$ represented the number of the neighboring voxels.}
    \label{tab:ana_neigh}
    \setlength{\tabcolsep}{1.2mm}{
      \begin{tabular}{cccc}
        \toprule
         $K$  & 8  & 12  & 16  \cr
        \midrule
         CD [m] $\downarrow$  & 0.431  & 0.685  & \textbf{0.425}  \\
         F-Score $\uparrow$   & 7.591  & 7.691  & \textbf{8.021}  \\
        \bottomrule
      \end{tabular}
      }
\end{table}

\textbf{Sensitivity of Neighboring Voxels.}
In our point-voxel interaction module, we search for the neighboring voxels of each upsampled point to improve the points' perception capability of their surroundings. To explore the impact of the number of neighboring voxels, we conduct a detailed analysis as shown in Table \ref{tab:ana_neigh}. We can see that when the number of neighboring voxels $K$ is set to 16, we achieve optimal results across both the CD and F-Score metrics.

\renewcommand{\arraystretch}{1.5}
\begin{table}
    \centering
    \fontsize{9}{9}\selectfont
    \caption{Analysis of the parameter $\lambda$ in our loss function.}
    \label{tab:ana_param}
    \setlength{\tabcolsep}{1.2mm}{
      \begin{tabular}{cccccc}
        \toprule
         $\lambda$  & 0.0  & 1.0 & 2.0  & 3.0  & 4.0   \cr
        \midrule
         CD [m] $\downarrow$  & 0.428  & \textbf{0.425}  & 0.431  & 0.433  & 0.426  \\
         F-Score $\uparrow$   & 7.809  & \textbf{8.021}  & 7.616  & 7.505  & 7.642  \\
        \bottomrule
      \end{tabular}
      }
\end{table}

\begin{figure}
  \centering
  \includegraphics[width=1.0\linewidth]{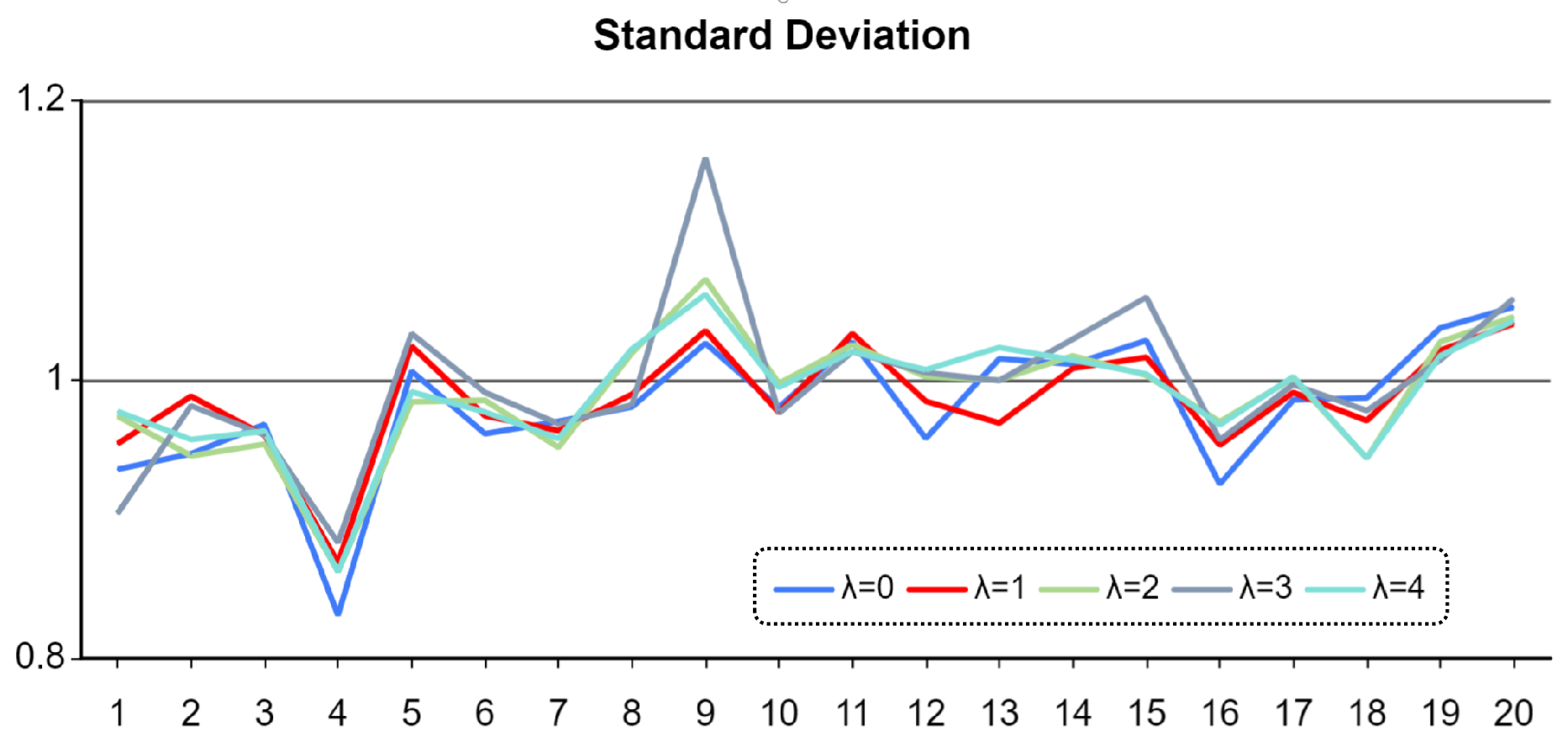}
  \caption{Standard deviation of the predicted noise over different regularization weights.}
  \label{fig:comp_lamdba}
\end{figure}

\textbf{Sensitivity of Noise Regularization Weight.}
In this work, we propose a noise regularization in the loss function to ensure that the predicted noise approximates the normal distribution for further improving the quality of generated points. In this section, we analyze the coefficient $\lambda$ to measure the effect of the regularization weight on the quality of the generated scene. 
The quantitative results are shown in Table \ref{tab:ana_param}. We can observe that the optimal performance is achieved when $\lambda$ is set to 1, while setting it to 0 leads to degraded model performance. As shown in Figure \ref{fig:comp_lamdba}, while $\lambda$ is set to 1.0, the corresponding standard deviation during training is closer to 1.0 compared to other regularization weights. 
Besides, when $\lambda$ exceeds 1.0, the metrics show a decline, and the corresponding curves of standard deviation fluctuate significantly around 1. This indicates that a larger regularization weight does not produce better results in our work. 

\renewcommand{\arraystretch}{1.5}
\begin{table}
    \centering
    \fontsize{9}{9} \selectfont
    \caption{Results with different voxel sizes.}
    \label{tab:voxel_size}
    \setlength{\tabcolsep}{1.2mm}{
      \begin{tabular}{ccc}
        \toprule
         Voxel size  & CD [m] $\downarrow$ & F-score $\uparrow$  \cr
        \midrule
         $128 \times 128\times 8$  & 0.458  & 6.704  \\
         $128\times 128\times 12$ & 0.491  & 4.442  \\
         $96\times 96\times 12$ & 0.671 & 5.964 \\
         $128\times 96\times 16$ & 0.435 & 7.526 \\
         $64\times 96\times 16$ & 0.694 & 7.734 \\
         $128\times 128\times 16$ & \textbf{0.425} & \textbf{8.021} \\
        \bottomrule
      \end{tabular}
      }
\end{table}

\textbf{Voxel Size.}
We investigate the impact of different voxel sizes on the upsampling performance. Specifically, we experiment with voxel sizes of $128 \times 128 \times 8 $, $ 128 \times 128 \times 12 $, $ 96 \times 96 \times 12 $, $ 128 \times 96 \times 16 $, and $ 64 \times 96 \times 16 $. As shown in Table \ref{tab:voxel_size}, the model achieves the best performance when the voxel size is set to $ 128 \times 128 \times 16 $. We attribute this to the fact that this configuration provides the highest number of voxels, enabling each point to effectively perceive sufficient contextual information from its surrounding environment. However, further increasing the voxel number would significantly increase computational costs. Therefore, we ultimately selected $ 128 \times 128 \times 16 $ as the voxel size for our method.

\section{Conclusion}
\label{sec:conclusion}
In this paper, we have proposed a novel diffusion model-based point upsampling network tailored for outdoor scenes. We first designed an effective voxel completion module to complete the initial voxels and enhance their feature representation. Then, we proposed a point-voxel interaction module that fully integrates the features from the upsampled points and voxels to achieve distinctive features. Moreover, we adopted the classifier-free guidance-based DDPMs to guide the generation of upsampled points. Finally, we conducted extensive experiments on benchmarks and provided detailed analysis. The quantitative and qualitative results demonstrate that our method achieves state-of-the-art performance. We believe that our work broadens the application of diffusion models and provides valuable insights and support for future outdoor scene upsampling and 3D scene understanding tasks.

\bibliographystyle{IEEEtran}

\begin{IEEEbiography}
  [{\includegraphics[width=1in,height=1.25in,clip,keepaspectratio]{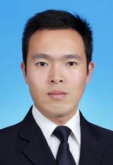}}]{Xianjing Cheng} 
  received the doctoral degree in the School of Computer Science and Technology from Guizhou University in 2022. He is a postdoctoral researcher at the School of Computer Science and Technology, Harbin Institute of Technology (Shenzhen). His research areas include stereo matching, image processing, 3D point cloud processing, scene semantic completion, and 3D object detection.
\end{IEEEbiography}

\begin{IEEEbiography}
  [{\includegraphics[width=1in,height=1.25in,clip,keepaspectratio]{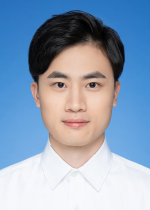}}]{Lintai Wu} received the B.S. degree in Management Information Systems from Wuhan University of 
  Technology in 2017, the M.S. degree in Computer Science and Technology from Harbin Institute of Technology, and the Ph.D. degree in Computer Science and Technology from Harbin Institute of Technology and City University of Hong Kong in 2025. He currently works as a Distinguished Associate Researcher at the College of Engineering, Huaqiao University. His research interests include 3D point cloud analysis and computer graphics.
\end{IEEEbiography}

\begin{IEEEbiography}
[{\includegraphics[width=1in,height=1.25in,clip,keepaspectratio]{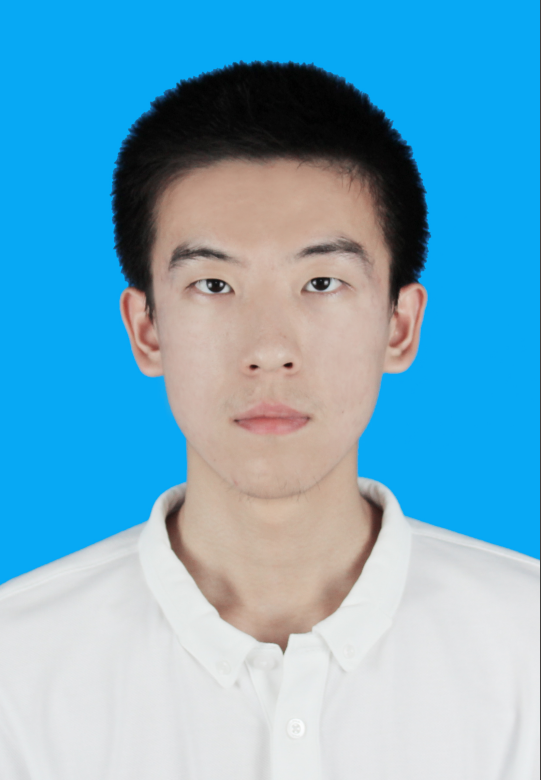}}]{Zuowen Wang} 
received the bachelor's degree in the School of Computer Science and Technology from Harbin Institute of Technology, Shenzhen in 2024. He is currently pursuing a M.S. degree with the School of Computer Science and Technology, Harbin Institute of Technology, Shenzhen. His research areas include 3D point cloud processing, scene semantic completion, and 3D occupancy prediction.
\end{IEEEbiography}

\begin{IEEEbiography}[{\includegraphics[width=1in,height=1.25in,clip,keepaspectratio]{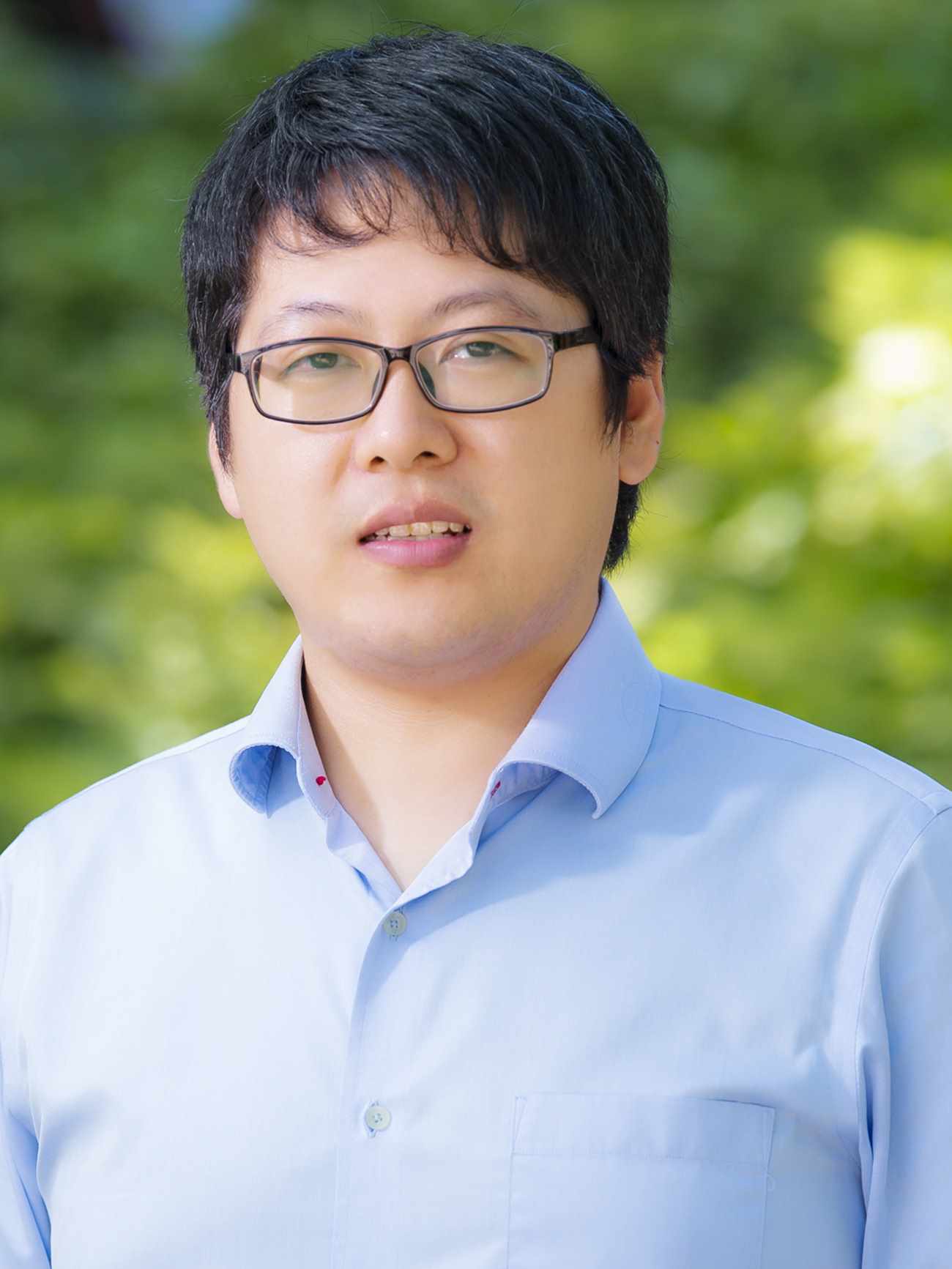}}]{Junhui Hou} (Senior Member, IEEE)
  is an Associate Professor with the Department of Computer Science, City University of Hong Kong.  His research interests are multi-dimensional visual computing. 

Prof. Hou received the Early Career Award (3/381) from the Hong Kong Research Grants Council in 2018 and the NSFC Excellent Young Scientists Fund in 2024. He has served or is serving as an Associate Editor for \textit{IEEE Transactions on Visualization and Computer Graphics}, \textit{IEEE Transactions on Image Processing}, \textit{IEEE Transactions on Multimedia}, and \textit{IEEE Transactions on Circuits and Systems for Video Technology}.
\end{IEEEbiography}

\begin{IEEEbiography}[{\includegraphics[width=1in,height=1.25in,clip,keepaspectratio]{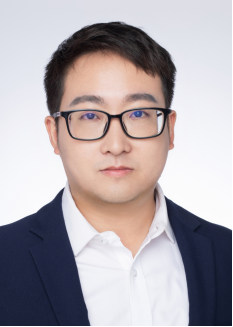}}]{Jie Wen} (Senior Member, IEEE)  received the Ph.D. degree in Computer Science and Technology at Harbin Institute of Technology, Shenzhen in 2019. 
  He is currently an Associate Professor at the School of 
  Computer Science and Technology, Harbin Institute of 
  Technology, Shenzhen. His research interests include 
  image and video enhancement, pattern recognition, and 
  machine learning. 
  He serves as an \textbf{Associate Editor} of \textit{IEEE Transactions on Image Processing}, \textit{IEEE Transactions on Information Forensics and Security},  \textit{Pattern Recognition}, and \textit{International Journal of Image and Graphics}, an \textbf{Area Editor} of \textit{Information Fusion}. 
  He also served as the \textbf{Area Chair} of \textit{ACM MM} and \textit{ICML}. 
\end{IEEEbiography}

\begin{IEEEbiography}
[{\includegraphics[width=1in,height=1.25in,clip,keepaspectratio]{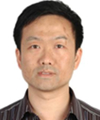}}]{Yong Xu} (Senior Member, IEEE)
was born in Sichuan, China, in 1972. He received the B.S. and M.S. degrees in 1994 and 1997, respectively, and the Ph.D. degree in pattern recognition and intelligence system from the Nanjing University of Science and Technology, China, in 2005. He is currently with the Shenzhen Graduate School, Harbin Institute of Technology. His current interests include pattern recognition, biometrics, machine learning, and video analysis.
\end{IEEEbiography}

\vfill

\end{document}